\documentclass[journal]{IEEEtran}
\usepackage{cite}
\usepackage{amsmath,amssymb,amsfonts}
\usepackage{mathtools}
\usepackage{bm}
\usepackage{graphicx}
\usepackage{textcomp}
\usepackage{xcolor,color}

\usepackage{amsthm}
\usepackage{amsmath}
\usepackage[ruled,vlined]{algorithm2e}
\usepackage[top=0.75in, left=0.75in, right=0.75in, bottom=0.8in]{geometry}
\usepackage{hyperref}
\usepackage{tikz}
\usetikzlibrary{positioning,automata}
\usepackage{subcaption}
\usepackage{empheq}
\usepackage{listings}
\usepackage{mathrsfs}
\usepackage{balance}

\newtheorem{proposition}{Proposition}
\newtheorem{remark}{Remark}

\newtheorem{example}{Example}

\definecolor{codegreen}{rgb}{0,0.6,0}
\definecolor{codegray}{rgb}{0.5,0.5,0.5}
\definecolor{codepurple}{rgb}{0.58,0,0.82}
\definecolor{backcolour}{rgb}{0.95,0.95,0.92}

\lstdefinestyle{mystyle}{
    backgroundcolor=\color{backcolour},   
    commentstyle=\color{codegreen},
    keywordstyle=\color{magenta},
    numberstyle=\tiny\color{codegray},
    stringstyle=\color{codepurple},
    basicstyle=\ttfamily\footnotesize,
    breakatwhitespace=false,         
    breaklines=true,                 
    captionpos=b,                    
    keepspaces=true,                 
    numbersep=5pt,                  
    showspaces=false,                
    showstringspaces=false,
    showtabs=false,                  
    tabsize=2
}
\newcommand{\q}{\mathbf{q}}
\renewcommand{\v}{\mathbf{v}}
\newcommand{\x}{\mathbf{x}}
\newcommand{\M}{\mathbf{M}}
\renewcommand{\k}{\mathbf{k}}
\newcommand{\J}{\mathbf{J}}
\newcommand{\G}{\mathbf{G}}
\newcommand{\bgamma}{\bm{\gamma}}
\newcommand{\blambda}{\bm{\lambda}}
\newcommand{\N}{\mathbf{N}}
\newcommand{\f}{\mathbf{f}}
\newcommand{\btau}{\bm{\tau}}
\newcommand{\z}{\mathbf{z}}

\newcommand{\p}{\mathbf{p}}
\renewcommand{\H}{\mathbf{H}}
\newcommand{\g}{\mathbf{g}}
\newcommand{\A}{\mathbf{A}}

\newcommand{\defeq}{\stackrel{\mathrm{def}}{=}}

\newcommand{\vf}[1]{{\bm{#1}}} 
\newcommand{\mf}[1]{{\mathbf{#1}}}

\DeclareMathOperator*{\argmin}{arg\,min}

\newcommand{\hl}[1]{#1}   
\newcommand{\hll}[1]{#1}  

\begin{document}

\title{\LARGE \bf CENIC: Convex Error-controlled Numerical Integration for Contact}

\author{Vince Kurtz and Alejandro Castro}

\maketitle
\thispagestyle{empty}

\begin{abstract}
State-of-the-art robotics simulators operate in discrete time. This requires
users to choose a time step, which is both critical and challenging: large steps
can produce non-physical artifacts, while small steps force the simulation to
run slowly. Continuous-time error-controlled integration avoids such issues by
automatically adjusting the time step to achieve a desired accuracy. But
existing error-controlled integrators struggle with the stiff dynamics of
contact, and cannot meet the speed and scalability requirements of modern
robotics workflows. We introduce CENIC, a new \hll{variable-step} integrator
that brings together recent advances in convex time-stepping and
error-controlled integration, inheriting benefits from both continuous
integration and discrete time-stepping. CENIC runs at fast real-time rates
comparable to discrete-time robotics simulators like MuJoCo, Drake and Isaac
Sim, while also providing guarantees on accuracy and convergence.
\end{abstract}

\begin{IEEEkeywords}
Simulation and Animation, Contact Modeling, Dexterous Manipulation, Dynamics.
\end{IEEEkeywords}

\section{Introduction}\label{sec:intro}

Simulation for contact-rich robotics has advanced significantly in recent years,
driven by demands in policy learning, data generation, and model-based control.
These efforts have led to highly optimized simulators that prioritize
throughput, with implementations targeting modern CPU and GPU architectures.
Most of these simulators---including Bullet \cite{bib:bullet}, MuJoCo
\cite{todorov2012mujoco}, Isaac Sim \cite{physx}, RaiSim \cite{bib:raisim},
Genesis \cite{Genesis}, and Drake \cite{drake}---rely on discrete time-stepping
with constraint-based contact formulations, a paradigm originally developed for
computer graphics and game physics, where speed is favored over accuracy. 

\begin{figure}
    \centering
    \includegraphics[width=0.9\linewidth]{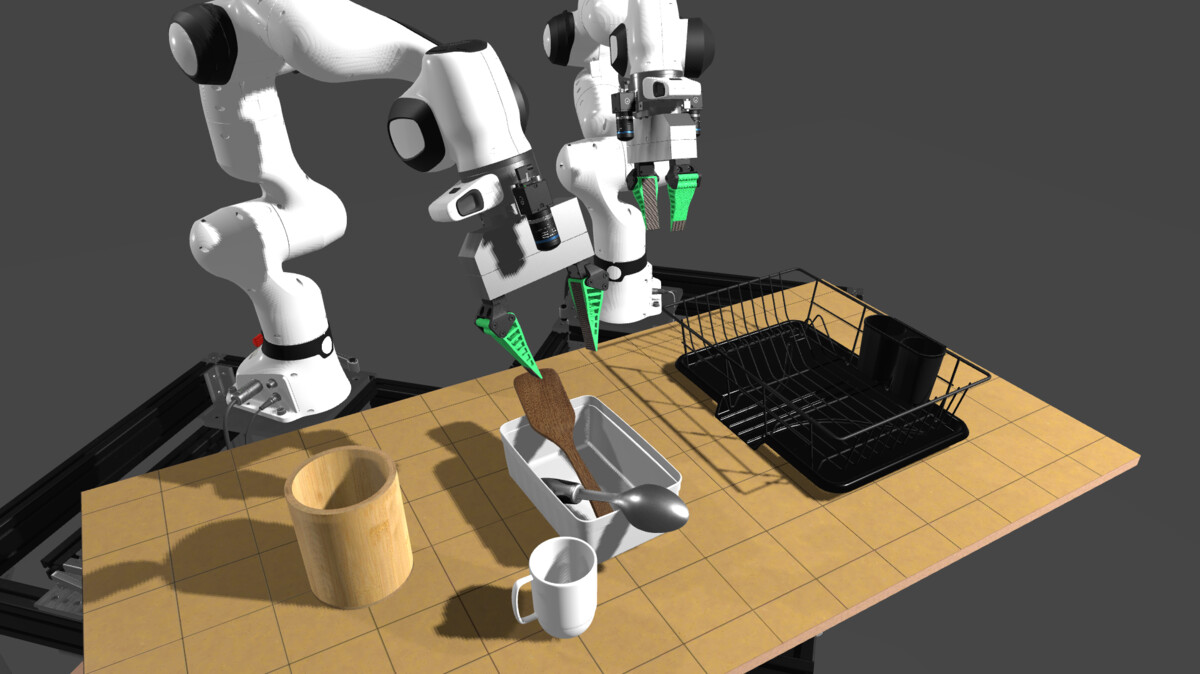}
    
    \vspace{0.3em}

    \includegraphics[width=0.9\linewidth]{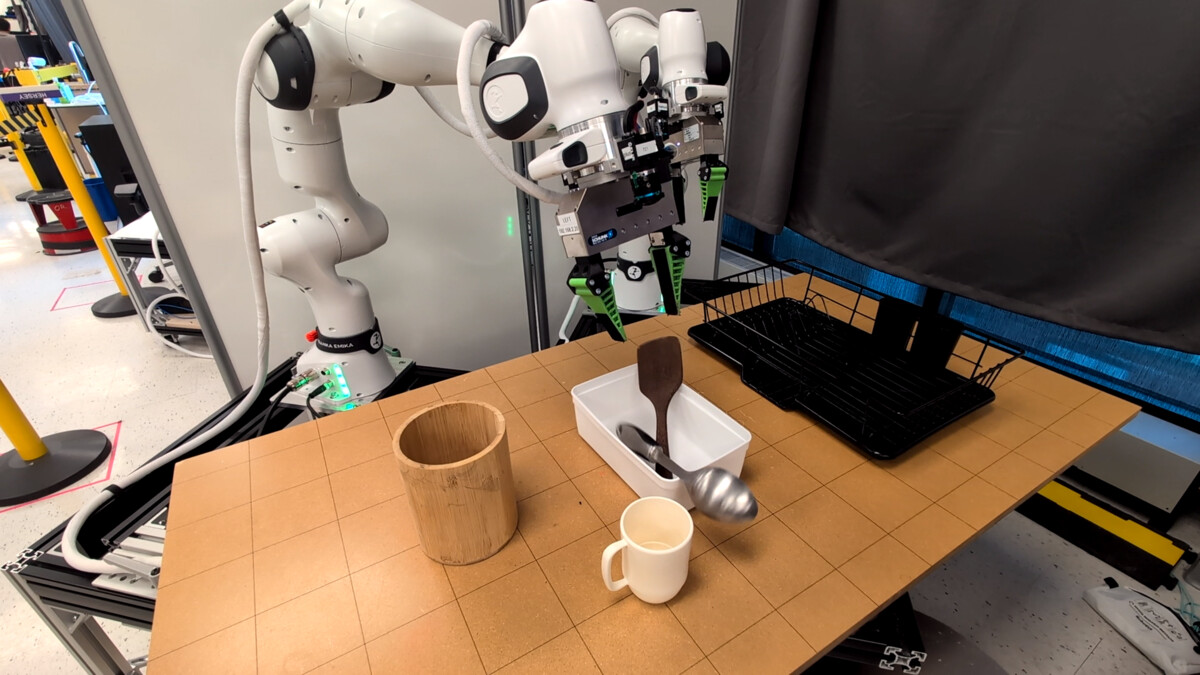}

    \caption{A challenging bimanual manipulation scenario in simulation (top)
    and on hardware (bottom). A teleoperated robot drops several objects
    (spatula, spoon, mug) into a thin-walled bin. It then dumps the contents of
    the bin on a dish rack composed of thin wires. This scenario is particularly
    challenging: complex geometry, thin objects, heavy arms, and stiff joint
    controllers create a wide range of time scales that pose significant
    numerical difficulties. Discrete time-stepping methods struggle with
    passthrough, rattling, and unstable contact forces; CENIC eliminates these
    artifacts while completing the simulation at a real-time rate over 300\%. }
    \label{fig:hero}
\end{figure}

A central limitation of discrete time-stepping lies in time step selection.
The appropriate time step depends on numerous factors, including object sizes,
contact stiffnesses, control frequencies, and geometric resolution. In dexterous
manipulation scenarios like that shown in Fig.~\ref{fig:hero}, large robotic
arms interacting with thin or stiff components introduce multiple time scales
that are difficult to resolve uniformly. Small steps are often needed to avoid
artifacts like interpenetration, tunneling, and impulse jitter. As simulation
scenarios become more complex---modeling full factory floors or domestic
environments---the number of contacts and degrees of freedom grows, making small
time steps increasingly inefficient. Additionally, unexpected events like
dropped objects or failed grasps can introduce unforeseen faster time scales,
leading to nonphysical artifacts and undermining the predictive validity of the
simulation.

In contrast, simulation tools in aerospace, automotive, and mechanical
engineering have long prioritized accuracy, supported by formal frameworks for
validation and verification (V\&V) and uncertainty quantification (UQ)
\cite{roache1998verification,oberkampf2010verification}.
Commercial multibody dynamics tools such as MSC Adams \cite{bib:HexagonAdams},
Siemens Simcenter Motion \cite{bib:SiemensSimcenter3d}, and RecurDyn
\cite{bib:EnginSoftRecurdyn} have been refined over decades and support
integration with finite element analysis (FEA), hydraulic and electronic
subsystem simulations \cite{nagel1975spice2}, and computer aided design (CAD).
These tools adopt a fundamentally different strategy: models of compliant
contact with regularized friction allow all forces---including contact and
friction---to be expressed as smooth functions of state.  Importantly, these
continuous models are grounded in experimentally validated contact mechanics,
such as Hertz theory \cite{bib:johnson1987} for elastic forces and the Hunt \&
Crossley \cite{bib:hunt1975} model for dissipative effects. The resulting
systems of ordinary differential equations (ODEs) are solved using
error-controlled \hl{(variable-step)} integration \cite{hairer1996solving}. Such
integrators adapt the time step to satisfy user-defined accuracy requirements.
This decouples numerical error from modeling error---critical for certified
engineering workflows.

But despite this reliability, engineering-level simulation tools are largely not
suitable for robotics. Applications like manipulation and locomotion require
high stiffnesses and tight regularization of friction to approximate rigid
contact. This leads to extremely stiff dynamics, where even state-of-the-art
\mbox{$L$-stable} stiff integrators \cite[\S IV.3]{hairer1996solving}---which
effectively damp high-frequency modes---are forced to take prohibitively small
steps, resulting in very slow simulations. Meanwhile, robotics applications like
reinforcement learning, interactive teleoperation, and model predictive control
require simulation speeds at (or well above) real time rates. 

\subsection{Contributions}

We introduce CENIC, the first error-controlled integrator for contact-rich
multibody dynamics tailored to robotics. CENIC builds on recent advances in
convex-optimization-based time-stepping
\cite{castro2022unconstrained,castro2024irrotational} to produce a convex
error-controlled integration scheme that automatically adjusts the time step to
meet a user-specified accuracy. Unlike fixed-step schemes, CENIC evolves the
simulated system in continuous time, mirroring the behavior of the physical
world.

CENIC combines well-established error control strategies
\cite{hairer1996solving} with recently-developed irrotational contact field
(ICF) theory \cite{castro2024irrotational}. As a result, CENIC blends the
performance of discrete time-stepping with the mathematical rigor of
error-controlled integration. This offers several benefits:

\begin{enumerate}
    \item \textbf{Accuracy}: Users specify a desired accuracy (e.g., the number
    of significant digits in the solution) rather than a fixed time step. CENIC
    automatically adjusts the step size to meet this accuracy requirement.

    \item \textbf{Consistency}: As accuracy is tightened, the numerical solution
    converges to the true continuous-time trajectory, avoiding artifacts such as
    the ``gliding'' characteristic of convex formulations, or the contact
    impulse ``jitter'' often seen in discrete time-stepping.

    \item \textbf{Validated contact models}. CENIC is built on engineering-grade
    continuous-time contact models.

    \item \textbf{Friction}. Unlike discrete-time schemes typically used in
    robotics, CENIC can rigorously model both static and dynamic friction\hl{,
    with guaranteed convergence.}

    \item \textbf{Modularity}: \hl{Controllers and external
    systems can be embedded in the convex formulation}, enabling stable
    simulation of complex user-defined components.

    \item \textbf{Convergence Guarantees}: Because the underlying time-stepping
    problem is convex, CENIC provides convergence guarantees independent of
    the current time step. This enables large steps even through complex contact
    transitions, where standard implicit integrators often fail or require many
    small steps.

    \item \textbf{Efficiency}: CENIC outperforms traditional
    error-controlled integrators by orders of magnitude. While error-estimation
    introduces some overhead, overall simulation times can be faster than
    discrete-time methods, thanks to the ability to take small steps only when
    necessary. We also demonstrate practical techniques---such as selective
    Hessian reuse and adaptive convergence tolerances---to further improve
    speed.
    
\end{enumerate}    

CENIC provides a new foundation for scalable, physically consistent
simulation in robotics, capable of handling large-scale, contact-rich
environments with provable guarantees on accuracy and convergence---ultimately
enabling tighter sim-to-real alignment for manipulation, locomotion,
reinforcement learning, and control applications.

\subsection{Organization}

Section~\ref{sec:background} reviews background and related work on robotics
simulation and error-controlled integration. Section~\ref{sec:model} introduces
our continuous-time modeling approach. We revisit ICF in Section~\ref{sec:icf},
highlighting its role as a foundation for CENIC. Section~\ref{sec:main} presents
the proposed CENIC integrator, followed by performance optimizations
(Section~\ref{sec:performance}), experimental results
(Section~\ref{sec:experiments}), discussion (Section~\ref{sec:discussion}) and
conclusions (Section~\ref{sec:conclusion}).
\section{Background}\label{sec:background}

Here we introduce notation (\ref{sec:background:modeling}) and review contact
models (\ref{sec:background:contact}). We then note that simulation approaches
can be broadly classified into continuous-time\hl{/variable-step}
(\ref{sec:background:continuous_formulations}) and discrete-time\hl{/fixed-step}
(\ref{sec:background:discrete_formulations}) formulations. The latter includes
convex approximations (\ref{sec:background:convex_approximations}), which offer
tractable formulations with strong theoretical guarantees.

\subsection{Multibody Dynamics}\label{sec:background:modeling}

We consider articulated rigid-body systems described by joint coordinates. The
state $\x = [\q; \v]$ collects generalized positions $\q \in \mathbb{R}^{n_q}$
and velocities $\v \in \mathbb{R}^{n_v}$, which are related kinematically by
$\dot{\q} = \N(\q)\v$. The dynamics under holonomic constraints
are given by
\begin{subequations}\label{eq:momentum_continuous}
\begin{align}
    &\M(\q)\dot{\v} + \k(\q, \v) = \btau + \J(\q)^T\f\hl{(\x)} + \G(\q)^T\blambda, \\
    &\text{s.t.} \quad \mf{c}(\q) = \mf{0},
    \label{eq:holonomic_constraints}
\end{align}
\end{subequations}
where $\M(\q)$ is the mass matrix and $\k(\q,\v)$ collects Coriolis and
gravitational terms. Lagrange multipliers $\blambda$ enforce holonomic
constraints $\mf{c}(\q) = \mf{0}$, with $\G(\q)=\partial_q\mf{c}\mf{N}(\q)$.
Non-holonomic constraints can be treated analogously, but since their main use
case---rolling---is already captured by our contact model, we focus on holonomic
constraints for clarity\footnote{\hl{While many physics engines categorize
constraints as \emph{bilateral} and \emph{unilateral}, the distinction between
\emph{holonomic} ($\mf{c}(\q) = 0$) and \emph{non-holonomic} ($\mf{c}(\q, \v) =
0$) is more useful in our context.}}. External forces, including actuation, are
grouped in $\btau$ and padded with zeros for unactuated degrees of freedom. The
block rows $\J_i(\q)\in\mathbb{R}^{3\times n_v}$ of the contact Jacobian define
the relative contact velocities $\vf{v}_{c,i}=\J_i(\q)\v$ at each contact. With
contact normal $\hat{\vf{n}}_i$, we define normal
$v_{n,i}=\vf{v}_{c,i}\cdot\hat{\vf{n}}_i$ and tangential
$\vf{v}_{t,i}=\vf{v}_{c,i}-v_{n,i}\hat{\vf{n}}_i$ velocities. Contact forces
$\vf{f}_i$ are concatenated as $\f=[\vf{f}_0; \vf{f}_1; \dots]$. \hl{We model
these forces as a function of state $\x$ via a compliant contact model.}

\subsection{Contact Modeling}\label{sec:background:contact}

When two solids make contact, they deform---at least microscopically---to
prevent interpenetration. These deformations generate stresses. Contact forces
are the integrated result of these stresses over the contact surface.

\textbf{Compliant contact} \hl{considers contact forces to be algebraic
functions of state. This captures quasistatic effects without solving a full
elasticity problem.} Compliant contact modeling dates back to the foundational
work of Hertz \cite{bib:hertz1881contact, bib:johnson1987}, describing elastic
contact under small strains. Hertz models are used in \emph{point contact}
approximations, which neglect the contact patch size. Later extensions
incorporate local geometry and volume estimates \cite{bib:gonthier2007,
bib:luo2006}.

When the contact patch size cannot be neglected, the elastic foundation model
(EFM) \cite{bib:johnson1987} offers a computationally efficient approximation.
EFM has been applied to meshed geometries in \cite{bib:hippmann2004,
sherman2011simbody}, and further refined in \cite{elandt2019pressure,
masterjohn2022velocity} to provide smooth, differentiable forces with coarse
meshes. Dissipative effects can be incorporated, most commonly using the Hunt \&
Crossley \cite{bib:hunt1975} model or similar variants.

Importantly, many of these compliant contact models have extensive experimental
validation, and are thus widely used in commercial multibody simulation tools
designed for certified engineering workflows \cite{bib:HexagonAdams,
bib:SiemensSimcenter3d, bib:EnginSoftRecurdyn}. However, for hard materials
such as metals or ceramics, the resulting dynamics become numerically stiff and
difficult to simulate. 

\textbf{Rigid contact} is a popular alternative to stiff compliant contact,
and idealizes bodies as infinitely stiff. This approximation underlies many
real-time physics and gaming engines and is discussed further in
Section~\ref{sec:background:discrete_formulations}.

\textbf{Friction} arises from complex surface interactions at the microscale,
including deformation of asperities and atomic adhesion. The Coulomb friction
model provides a simple and widely-used approximation of the resulting net
effects.

When contact surfaces do not slip, the Coulomb model specifies that friction
forces satisfy $\|\vf{f}_t\| \le \mu_s f_n$, where $f_n$ is the normal force,
$\vf{f}_t$ is the tangential component of the friction force, and $\mu_s$ is the
\emph{coefficient of static friction}. During slip, the maximum dissipation
principle (MDP) specifies that friction opposes the direction of slip,
\[
\vf{f}_t = -\mu_d \frac{\vf{v}_t}{\|\vf{v}_t\|} f_n,
\]
where $\mu_d \le \mu_s$ is the \emph{coefficient of dynamic friction}.

Coulomb's friction law is discontinuous and multivalued. That is, in
stiction, friction forces can have any magnitude between zero and $\mu_s f_n$.
To avoid this, \textbf{regularized friction} models friction as a continuous
function of slip velocity
\begin{equation}
    \vf{f}_t = -\mu(\Vert\vf{v}_t\Vert)\frac{\vf{v}_t}{\Vert\vf{v}_t\Vert}f_n,
    \label{eq:regularized_friction}
\end{equation}
where $\mu(\cdot)$ smoothly transitions between the static and dynamic
regimes (Fig. \ref{fig:dynamic_friction}). For instance, Simbody
\cite{sherman2011simbody} uses piecewise polynomials with $C^2\text{-smooth}$
transitions, with similar models in Adams \cite{bib:HexagonAdams} and RecurDyn
\cite{bib:schuderer2025}.

\subsection{Continuous-Time Contact Simulation}
\label{sec:background:continuous_formulations}

Together, compliant contact and regularized friction make the contact forces
$\vf{f}_i(\mf{x})$ continuous functions of the state. In the absence of
constraints, incorporating these forces into the dynamics
\eqref{eq:momentum_continuous} yields a system of ODEs. For clarity, we defer
discussing \hl{other} constraints to Sec.~\ref{sec:model:full}.

Continuous formulations offer several advantages for modeling contact dynamics:
they handle both static and dynamic friction, build on contact models validated
over decades of experimentation, and can leverage \hl{variable-step}
error-controlled integration, which adjusts the time step to bound numerical
deviations from true continuous solutions.

Error-controlled integration is a well-established topic, see
\cite{hairer1996solving} for an excellent reference. \hll{While the true
continuous-time solution is rarely available in closed form, we can estimate the
error by comparing a higher-order solution to a lower-order one.
Error-controlled integration adapts the time step to keep this estimated error
below a user-specified accuracy tolerance. Specifically,} given an ODE
\begin{equation*}
    \dot{\x} = \mf{f}(t, \x),
\end{equation*}
the integrator attempts a step of size $\delta t$  from time $t^n$ to
$t^{n+1}=t^n+\delta t$, where superscripts index the step. Within this step, it
computes two estimates of the next state, \(\x^{n+1}\) and \(\hat{\x}^{n+1}\), and
uses their difference to approximate the local truncation error
\begin{equation}\label{eq:one_step_error}
    e^{n+1} = \|\x^{n+1} - \hat{\x}^{n+1}\| \approx \bar{c}\,\delta t^p,
\end{equation}
where \(p\) is the order of the error estimate and \(\bar{c}\) a constant. For
example, the Dormand-Prince integrator \cite{bib:dormand} computes \(\x^{n+1}\)
with order 5 and $\hat{\x}^{n+1}$ with order 4, leading to a fifth order error
estimate ($p=5$). From $e^{n+1} = \bar{c}\,\delta t^p$ and $\varepsilon_\text{acc} =
\bar{c}\,\delta t_\text{new}^p$, $\bar{c}$ is eliminated to estimate a step size
$\delta t_\text{new}$ that attains a user-specified accuracy
$\varepsilon_\text{acc}$,
\begin{equation}
    \delta t_\text{new} \gets \delta t \left(\frac{\varepsilon_\text{acc}}{e^{n+1}}\right)^{1/p}.
    \label{eq:step_size_udpate}
\end{equation}
If \(e^{n+1} < \varepsilon_\text{acc}\), the step \hll{is} accepted and the
state is advanced to $\x^{n+1}$. Otherwise, the step is rejected, the time step
is adjusted using \eqref{eq:step_size_udpate}, and the process is repeated until
the estimated error falls below the specified accuracy.

Continuous formulations with error control are the foundation of most commercial
multibody dynamics tools, including Adams \cite{bib:HexagonAdams}, RecurDyn
\cite{bib:EnginSoftRecurdyn}, Simcenter 3D \cite{bib:SiemensSimcenter3d},
Simpack \cite{bib:Simpack}, and MotionSolve \cite{bib:MotionSolve}. These tools
are widely used in engineering, where predictive accuracy and fidelity are
critical. Continuous simulation has also seen some success in biomechanics
simulation, with the open-source Simbody~\cite{sherman2011simbody} and
commercial Hyfydy~\cite{bib:Geijtenbeek2021Hyfydy} as notable examples.

However, large mass ratios, stiff contact dynamics, and tightly regularized
friction yield extremely stiff ODEs. As a result, even state-of-the-art
\mbox{$L$-stable} integrators often struggle or stall on practical robotics
problems. \hl{As a result, \hll{variable-step} methods are rarely used in
robotics and graphics applications, where faster simulation takes
\hll{precedence over accuracy}. A seemingly related work is asynchronous
contact mechanics \cite{harmon2009asynchronous, ainsley2012speculative}, an
event-driven method that maintains a separate time step for each contact force.
This is different from error-controlled integration and as such does not provide
accuracy guarantees. }

\subsection{Discrete-Time Contact Simulation}
\label{sec:background:discrete_formulations}

The stiffness of continuous-time dynamics makes accurate integration
computationally demanding. Consequently, modern robotics and real-time
simulators favor fixed-step discrete-time schemes, which forgo accuracy
guarantees but maintain predictable computational costs.

Discrete-time formulations also permit the rigid-contact approximation,
effectively modeling contact stiffness as infinite. Rigid contact with Coulomb
friction can produce Painlev\'e paradoxes \cite{bib:pfeiffer1996multibody},
where solutions may not exist. Theory resolves these paradoxes by allowing
discrete velocity jumps and impulsive forces, formally casting the problem as a
differential variational inequality (DVI) \cite{bib:pang2008differential}. Thus
discrete formulations integrate the dynamics \eqref{eq:momentum_continuous}
within the time interval $[t^n, t^{n+1})$ to derive a discrete momentum balance
where impulses $\bgamma_i = \int_{t^{n}}^{t^{n+1}} \vf{f}_i(t) \, dt$ replace
contact forces $\vf{f}_i$ and discrete velocity changes replace continuous
accelerations \cite{bib:stewart1996implicit, bib:anitescu1997}. Additional
constraints enforce Coulomb's law and the MDP, leading in general to a nonlinear
complementarity problem (NCP) where velocities, impulses and constraint
multipliers are the unknowns.

NCPs are NP-hard \cite{bib:Kaufman2008}, and are extremely difficult to
solve in practice. The literature on NCP-based simulation is vast, with
techniques including linearized friction cones \cite{bib:anitescu1997,
bib:stewart1998convergence}, projected Gauss-Seidel (PGS) solvers
\cite{bib:bullet,bib:erleben2007velocity, bib:raisim}, non-convex optimization
\cite{bib:Kaufman2008, bib:li2020ipc,lan2022affine,huang2025stiffgipc},
non-smooth formulations \cite{bib:macklin2019}, interior point methods
\cite{bib:howell2022dojo}, alternating direction method of multipliers (ADMM)
solvers \cite{carpentier2024compliant}, \hl{and augmented lagrangian methods
\cite{lee2025variations}. Many of these approaches have demonstrated excellent
performance in practice, thanks in large part to careful domain-specific
engineering in the NCP solver design. However, this comes at the expense of
sacrificing accuracy and convergence guarantees. }

Though not widely accepted in robotics, some commercial platforms such as
Algoryx \cite{bib:algoryx} and Vortex \cite{bib:vortex} also adopt discrete
time-stepping methods for real-time multibody dynamics, targeting domains like
virtual prototyping, operator training, and vehicle simulation.

\subsection{Convex Approximations}
\label{sec:background:convex_approximations}

Without existence and convergence guarantees, NCP-based simulations face
robustness challenges. Techniques such as constraint relaxation and numerical
stabilization (e.g. \emph{constraint force mixing} \cite{ode}) can alleviate
these challenges but introduce artificial compliance---the very phenomenon rigid
NCP formulations were designed to avoid. \hl{Additionally, non-convex methods
often fail to converge with large time steps, even when the underlying problem
is not a strict NCP \cite{castro2020transition}.}

To avoid these issues, Anitescu introduced a \textit{convex relaxation} of the
contact problem \cite{bib:anitescu2006}. Later, Todorov introduced a regularized
convex formulation with a unique solution in MuJoCo \cite{bib:todorov2014},
though at the expense of non-physical regularization parameters and noticeable
constraint drift \cite{bib:simbenchmark}. The implementation in MuJoCo is
remarkably performant and thus widely used in the robotics and RL communities.
Chrono \cite{mazhar2013chrono}, Siconos \cite{acary2007siconos}, and Drake
\cite{drake} also implement convex time-stepping methods. The SAP
\cite{castro2022unconstrained} formulation in Drake leverages numerical
regularizations that correspond to the physics of compliant contact.

However, all of these Anitescu-based convex relaxations suffer from common
problems. They all exhibit \emph{gliding} or \emph{hydroplaning} artifacts,
where sliding objects hover at a distance proportional to slip velocity. This
gliding artifact was reported inadequate for quadruped simulation in
\cite{bib:lelidec2024}. Even more acute, for compliant formulations like those
in MuJoCo \cite{bib:todorov2014} and SAP \cite{castro2022unconstrained}, gliding
is also proportional to the dissipation parameter and does not vanish as the
time step is reduced \cite{castro2024irrotational}. Thus users are forced to
trade-off contact modeling accuracy with numerical artifacts, often simulating
models that are further from reality to limit numerical issues.

Most recently, irrotational contact fields (ICF) theory
\cite{castro2024irrotational} establishes conditions under which
engineering-grade compliant contact models can be embedded in a convex
formulation. Moreover, ICF introduces a \emph{lagged} convex approximation that
eliminates gliding artifacts. ICF regularizes friction in terms of a single
\emph{stiction tolerance} velocity parameter $v_s$ such that friction forces are
given by
\begin{equation}
    \vf{f}_t = -\mu_d\frac{\vf{v}_t}{\sqrt{\Vert\vf{v}_t\Vert^2+v_s^2}}f_n,
    \label{eq:icf_regularized_friction}
\end{equation}
with magnitude
\begin{equation*}
    \hl{\Vert\vf{f}_t\Vert=\mu_dg(\Vert\vf{v}_t\Vert/v_s)f_n.}
\end{equation*}
Here \hl{$g(s)=s/\sqrt{s^2+1}$} is a sigmoid function resulting from ICF's
convex potential. Sticking contact corresponds to $\Vert\vf{v}_t\Vert < v_s$,
while slipping contacts with $\Vert\vf{v}_t\Vert > v_s$ satisfy Coulomb's law
and the MDP. ICF's robustness and strong mathematical guarantees allow it to
operate with extremely tight regularization ($v_s \approx 0.1\,\text{mm/s}$), as
required for demanding robotics applications like dexterous manipulation. As
reference, the biomechanics simulator Simbody \cite{sherman2011simbody} uses a
default stiction tolerance of $1\,\text{cm/s}$ while Hyfydy
\cite{bib:Geijtenbeek2021Hyfydy} and MSC Adams \cite[p.~145]{adamsUserGuide} use
$10\,\text{cm/s}$.

\section{Continuous-Time Dynamics Model}
\label{sec:model}

CENIC takes a continuous-time approach
(Section~\ref{sec:background:continuous_formulations}). Here we detail our
particular choices for modeling contact, friction, and external control systems
in continuous time.

\subsection{Compliant Contact}
\label{sec:model:compliant_contact}

We build on the contact models implemented in Drake
\cite{castro2024irrotational, masterjohn2022velocity}. These compliant models
provide algebraic formulae to compute the elastic contribution to the normal
contact force $f_{e,i}(\q)$ at the $i$-th contact point as a function of
configuration $\q$. For simplicity, in the following we drop the contact index
$i$.

\textbf{Point contact} models the normal force as $f_e(\q) = k\,(-\phi(\q))_+$,
where $\phi(\q)$ is the signed distance between two interacting geometries
(negative when geometries overlap),  $k$ is the stiffness, and $(a)_+ =
\max(0,a)$.

\textbf{Hydroelastic contact} \cite{elandt2019pressure} models continuous
contact patches, discretized into polygons \cite{masterjohn2022velocity} of area
$A(\q)$, with elastic normal force $f_e(\q)=A(\q)\,p_e(\q)$, where $p_e(\q)$ is
the hydroelastic contact pressure at the polygon's centroid
\cite{masterjohn2022velocity}. In this model, stiffness is parameterized by a
\emph{hydroelastic modulus} (similar to, though not the same as, Young's
modulus for a given material).

For both point and hydroelastic models, we incorporate the Hunt \& Crossley
\cite{bib:hunt1975} model of dissipation to write the normal force as a function
of the system state $\x$
\begin{equation}
	f_n(\x)=f_e(\q)(1-d\,v_n(\x))_+,
    \label{eq:compliant_normal_force}
\end{equation}
where $v_n(\x)$ is the normal velocity (positive when objects move apart) and
$d$ is the Hunt \& Crossley dissipation parameter.

\subsection{Regularized Friction}

\begin{figure}[!h]
	\centering
	\includegraphics[width=\linewidth]{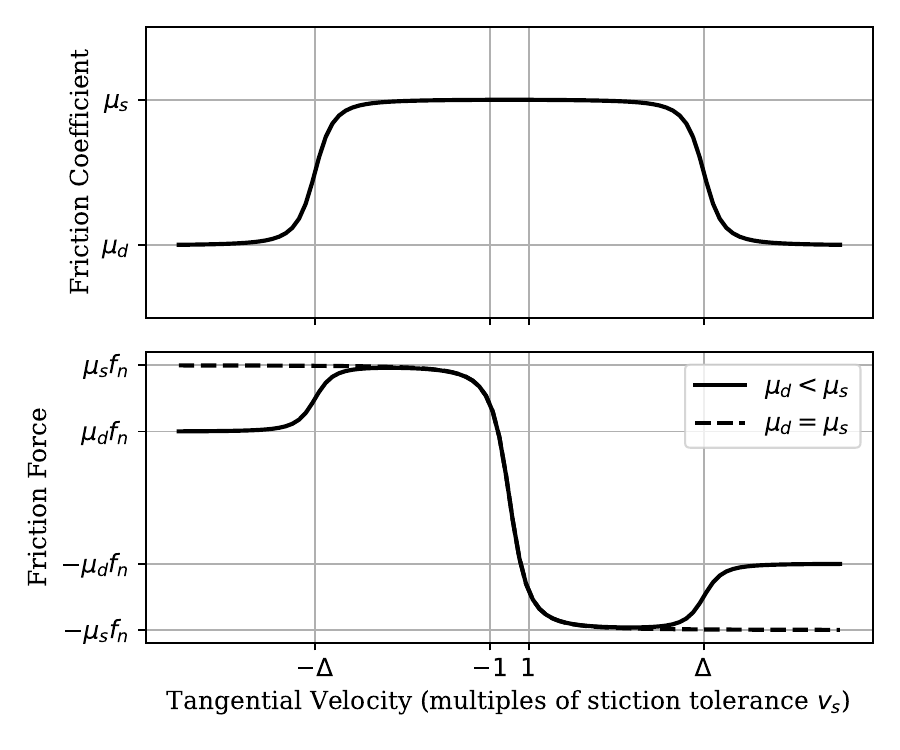}
	\caption{CENIC's regularized friction model captures both static ($\mu_s$)
		and dynamic ($\mu_d$) friction regimes. Our default stiction tolerance
		is $v_s = 0.1 \,\text{mm/s}$, with a transition width of $\Delta = 10$.}
	\label{fig:dynamic_friction}
\end{figure}

We design a regularized friction model that extends
\eqref{eq:icf_regularized_friction} to resolve both static and dynamic friction
\begin{equation}
	\vf{f}_t = -\mu(\Vert\vf{v}_t\Vert/v_s)\frac{\vf{v}_t}{\sqrt{\Vert\vf{v}_t\Vert^2+v_s^2}}f_n,
	\label{eq:cenic_regularized_friction}
\end{equation}
where the friction coefficient is given by
\begin{equation}\label{eq:dynamic_friction}
	\mu(s) = (\mu_s - \mu_d)\sigma(s) + \mu_d,
\end{equation}
with
\begin{equation*}
	\hl{\sigma(s) = \frac{1}{2} \left( 1 - \frac{g(|s|-\Delta)}{g(\Delta)} \right),}
\end{equation*}
conveniently defined in terms of the ICF's sigmoid function \hl{$g(\cdot)$} in
\eqref{eq:icf_regularized_friction}. $\Delta$ measures the width of the
transition region, as shown in Fig.~\ref{fig:dynamic_friction}. We found that
CENIC works particularly well with $\Delta=10$, while maintaining ICF's tight
stiction modeling when $\Vert\vf{v}_t\Vert < v_s$.


\subsection{External Systems}

\begin{figure}
	\centering
	\includegraphics[width=0.9\linewidth]{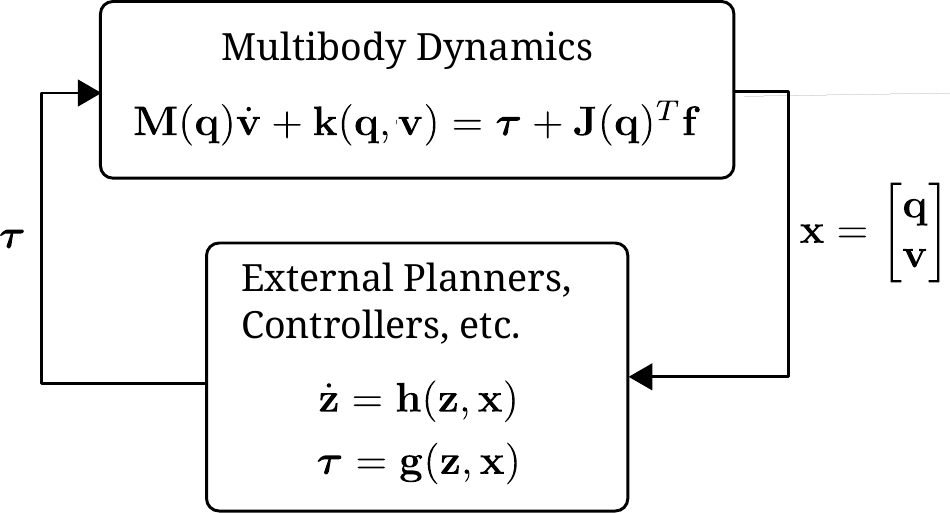}
	\caption{In practice, robotics simulations involve multibody dynamics
		connected to controllers, planners, and other components. We abstract
		these elements as a single external dynamical system. Unlike discrete
		formulations, CENIC integrates external systems implicitly, an effective
		strategy even for stiff feedback dynamics. }
	\label{fig:system_diagram}
\end{figure}

Useful robotics simulations involve not only multibody dynamics, but also
external systems like controllers, sensors, planners, and learned policies.
\hl{Such components are critical to simulate accurately, as roboticists often
use simulation to evaluate or improve these components.} Such external systems
can be described by a separate dynamical system in a feedback loop, as
illustrated in Fig.~\ref{fig:system_diagram}.

The external system takes as input the multibody state $\x$ and outputs actuator
forces $\btau$. These forces may depend on the external system's own state $\z$,
which evolves with dynamics
\begin{subequations}
	\begin{align}
		\dot{\z} & = \mf{h}(\z, \x),
		\label{eq:ext_sys_dynamics}\\
		\btau       & = \mf{g}(\z, \x),
		\label{eq:ext_sys_output}
	\end{align}
	\label{eq:external_system_dynamics}
\end{subequations}
\hl{where we assume that $\mf{h}$ and $\mf{g}$ are $C^1$ differentiable.}

The external system is often composed of many interconnected sub-systems. In
Drake, dynamical systems are modeled using \emph{block diagrams} with
\emph{graph-based representations}, where nodes (blocks) denote components such
as integrators, gains, or state-space models, and directed edges (signals)
define variable flow between them. MATLAB, SimuLink, and OpenModelica provide
similar representations.

\subsection{Full Dynamical System}\label{sec:model:full}

The constrained multibody dynamics \eqref{eq:momentum_continuous}, combined with
the compliant contact and regularized friction models introduced above, describe
the dynamics of the multibody system. Together with the external system dynamics
\eqref{eq:external_system_dynamics}, the entire system evolves according to
\begin{subequations}
	\begin{align}
	\M(\q)&\dot{\v} + \k(\q,\v)=\nonumber\\
	      &\quad\quad \mf{g}(\z,\x) + \J(\q)^{\!T}\f(\x) + \G(\q)^{\!T}\blambda, \label{eq:full_system_dynamics.dyn}\\
	&\dot{\q} = \N(\mf{q})\mf{v}, \label{eq:full_system_dynamics.kin}\\
	&\dot{\mf{z}} = \mf{h}(\z,\x), \label{eq:full_system_dynamics.z}\\
	\text{s.t.}\quad &\mf{c}(\q) = \mf{0}. \label{eq:full_system_dynamics.constr}
	\end{align}
	\label{eq:full_system_dynamics}
\end{subequations}

\hll{Note that contact forces $\f(\x)$ are a function of state via the compliant
(though often very stiff) model described above. These contact forces must vary
continously with state. Compliant point contact models with smooth geometries
(spheres, capules, etc.) and hydroelastic contact with any geometry
\cite{masterjohn2022velocity} satisfy this requirement.}

This index-3 system of differential-algebraic equations (DAEs) fully defines the
\emph{model}, encompassing both physics and external interactions. Its high
index introduces challenges such as constraint drift and difficulty in finding
consistent initial conditions. Moreover, \eqref{eq:full_system_dynamics} is
highly stiff, and even state-of-the-art stiff integrators
(Sec.~\ref{sec:experiments}) often struggle or fail to solve it. In the
following sections, we show how CENIC brings together the best of convex
time-stepping and error-controlled integration to efficiently solve this DAE.

\hl{
\subsection{Modeling versus Numerical Error}\label{sec:model:regularization}

All models approximate reality, and numerical methods, in turn, approximate the
solution of those models. Central to this work is the distinction between
\emph{modeling} and \emph{numerical} errors. Modeling error measures the
discrepancy between the simulated equations of motion
\eqref{eq:full_system_dynamics} and the true system dynamics, whereas numerical
error measures the accuracy with which those equations are solved. For a given
model, CENIC controls numerical error using established error-control techniques
\cite{hairer1996solving}.

Modeling choices remain the responsibility of the modeling engineer. Both rigid
and compliant contact approximate reality, and the appropriate model depends on
the application and computational requirements. Likewise, regularized friction
and compliant contact introduce parameters, such as stiction velocity and
contact stiffness, that must be selected or identified appropriately, much like
other physical parameters such as mass or inertia.

\hll{These modeling choices can also affect computational performance,
particularly through the numerical stiffness introduced by contact parameters.
While stiffer models may require smaller time steps with traditional
error-control schemes, CENIC builds on the highly stable ICF scheme to
efficiently handle stiff dynamics and frictional contact at larger step sizes.
Moreover, CENIC retains the benefits of error-controlled integration,
eliminating the need for manual time-step tuning.}}

\section{Irrotational Contact Fields}
\label{sec:icf}

Given its central role in this work, this section briefly introduces ICF
\cite{castro2024irrotational}, related notation, and the properties that make
ICF a key building block for CENIC. We refer the reader to
\cite{castro2024irrotational} for further details and derivations.

ICF advances the multibody state ($\x = [\q; \v]$) with a first-order
symplectic Euler scheme, where velocities $\v^{n+1}$ are solved implicitly from
the discrete momentum balance
\begin{equation}\label{eq:momentum_discrete}
    \M^n(\v^{n+1} - \v^n) + \delta t \k^n = {\J^n}^T \bgamma(\v^{n+1}; \x^n, \delta t),
\end{equation}
and then used to update configurations
\begin{equation}\label{eq:q_update}
	\q^{n+1} = \q^n + \delta t \N^n \v^{n+1}.
\end{equation}

The mass matrix ($\M^n = \M(\q^n)$), Coriolis and potential terms ($\k^n =
\k(\q^n, \v^n)$), constraint Jacobian ($\J^n = \J(\q^n)$), and projection matrix
($\N^n = \N(\q^n)$) are approximated explicitly to first order, much like a
first-order Implicit-Explicit (IMEX) scheme \cite{bib:ascher1995}. For
simplicity, we defer discussion of constraints and actuation to
Sections \ref{sec:icf:constraints} and \ref{sec:icf:linear_control}.

The semicolon notation $\bgamma(\v^{n+1}; \x^n, \delta t)$ denotes that
impulses, though evaluated implicitly on $\v^{n+1}$, stem from an
\emph{incremental potential} $\ell(\v; \x^n, \delta t)$ built \hl{using the
state at the beginning of the integration step $\x^n$}. More precisely,
$\bgamma(\v^{n+1}; \x^n, \delta t)$ concatenates impulses $\bgamma^{n+1}_i$ from
each constraint. 

\subsection{Compliance Approximation}
ICF approximates normal impulses \eqref{eq:compliant_normal_force} to
first-order, fully implicit in $\v^{n+1}$ to ensure strong numerical stability.
For normal contact velocity $v_{n, i}$, the corresponding impulse is
\begin{align}
    \gamma_{n,i}^{n+1}(v_{n,i};~& \x^n, \delta t)=\nonumber\\
	&\delta t\,\left(f_{e,i}^n-\delta t\,k_i\,v_{n,i}\right)_+\left(1-d_i\,v_{n,i}\right)_+,
    \label{eq:icf_normal_approximation}
\end{align}
where $f_{e,i}^n = f_{e,i}(\q^n)$ is the elastic contribution given
algebraically by the compliant contact model \eqref{eq:compliant_normal_force},
$k_i$ is the contact stiffness and $d_i$ the Hunt \& Crossley dissipation. For
point contact, stiffness is a user supplied parameter. For hydroelastic contact,
the stiffness combines cell area and hydroelastic pressure gradient
\cite{masterjohn2022velocity}.

\subsection{Friction Approximation}

ICF theory describes a family of convex contact approximations. \emph{SAP},
\emph{Lagged} and \emph{Similar} are solutions to ICF that belong to this family
\cite{castro2024irrotational}. CENIC builds on the \emph{Lagged} approach, which
approximates the tangential component of the impulse to first order as
\begin{equation}
    \bgamma^{n+1}_{t, i} = -\mu_{d,i}\frac{\vf{v}_{t,i}^{n+1}}{\sqrt{\Vert\vf{v}_{t,i}^{n+1}\Vert^2+v_s^2}}\gamma_{n,i}^n.
    \label{eq:icf_friction_approximation}
\end{equation}

Note that the tangential component is implicit in the velocities, while the
normal component is \emph{lagged}, with $\gamma_{n,i}^n=\delta t\,f_{n,i}(\x^n)$
\hl{evaluated explicitly at the beginning of the integration step
\eqref{eq:compliant_normal_force}}.

\subsection{Convex Formulation}

ICF shows we can solve an unconstrained convex optimization problem for
velocities at the next time step
\begin{equation}
		\v^{n+1} = \argmin_\v \ell(\v; \x^n, \delta t),
		\label{eq:sap_v}
\end{equation}
where optimality condition $\nabla \ell = 0$ corresponds to the discrete
momentum balance \eqref{eq:momentum_discrete}, together with the first-order
approximations \eqref{eq:icf_normal_approximation} and
\eqref{eq:icf_friction_approximation}.

The cost can be written as
\begin{equation}
    \ell(\v; \x^n, \delta t) = \frac{1}{2}\v^T\A\v -\mf{r}^T\v + \ell_c(\v; \x^n, \delta t),
    \label{eq:icf_cost}
\end{equation}
where $\A=\M^n$, $\mf{r} = \M^n\v^n-\delta t \k^n$, and $\ell_c=\sum_i \ell_i$
is separable into individual potentials $\ell_i$ for each constraint. The
gradient of the constraint potential satisfies
\begin{equation}\label{eq:constraint_cost_gradient}
    \nabla\ell_c(\v; \x^n, \delta t) = -{\J^n}^T \bgamma(\v; \x^n, \delta t).
\end{equation}
In addition to contact, constraints such as couplers, distance constraints,
limits, and welds can be included in this formulation, as described below. 

\hl{
\begin{remark}
    In the absence of constraints, ICF reduces to a simple semi-implicit Euler
    scheme,
    \begin{gather*}
        \v^{n+1} = \v^n - \delta t \M^{-1}(\q^n)\k(\q^n, \v^n),\\
        \q^{n+1} = \q^n + \delta t \N(\q^n)\v^{n+1}.
    \end{gather*}
    The convex cost \eqref{eq:icf_cost} becomes a simple quadratic, which is
    optimized with a single linear solve. 
\end{remark}
}

\subsection{Limit and Holonomic Constraints}\label{sec:icf:constraints}

ICF can incorporate holonomic constraints \eqref{eq:holonomic_constraints} and
limit constraints, which we express in the general form
\begin{equation*}
    c_{l,i} \le c_i(\q) \le c_{u,i},
\end{equation*}
for the $i\text{-th}$ constraint, with $c_{l,i}$ and $c_{u,i}$ the lower and
upper bounds. The holonomic case corresponds to $c_{l,i}=c_{u,i}$.

Limit constraints admit impacts with discontinuous velocity jumps and thus must
be treated within the context of DVIs, where complementarity conditions between
constraints and their multipliers are enforced. ICF avoids the complementarity
formulation via a regularized (compliant) model
\begin{align*}
    \lambda_{l,i} &= \phantom{-}\hl{\kappa}_i(c_{l,i} - c_i - \hl{\eta}_i \dot{c}_i)_+H(c_{l,i} - c_i),\\
    \lambda_{u,i} &= -\hl{\kappa}_i(c_i - c_{u,i} + \hl{\eta}_i \dot{c}_i)_+H(c_i - c_{u,i}),
\end{align*}
where the Heaviside function $H(x)$ is necessary to avoid action at a distance.
Stiffness and Kelvin--Voigt damping are set according to the
\emph{near-rigid} estimation \cite{castro2022unconstrained}
\begin{align}
    \hl{\kappa}_i = \frac{1}{4\pi^2\beta^2}\frac{\textrm{m}_i}{\delta t^2},\quad
    \hl{\eta}_i = \frac{\beta}{\pi}\delta t,
    \label{eq:near_rigid}
\end{align}
with $\textrm{m}_i=\Vert\mf{W}_{ii}\Vert^{-1}$, and $\mf{W}_{ii}$ is a diagonal
estimate of the Delasus operator
$\mf{W}=\mf{G}\mf{M}\mf{G}^T$\cite{castro2022unconstrained}. In this
\emph{near-rigid} regime, the $i\text{-th}$ constraint behaves as a critically
damped harmonic oscillator with period $\beta\delta t$. A value $\beta=0.1$ is
typical and delivers a tight constraint enforcement in practice. \hll{Note that
the effective stiffness scales with $\delta t^{-2}$, so the rigid
approximation becomes more accurate as error-control reduces the time step.}

ICF uses a first-order Taylor expansion
\begin{align*}
    c_i^{n+1} &= c_i^n + \delta t\dot{c}_i^{n+1},\\
    \dot{c}_i^{n+1} &= \mf{G}_i^n\v^{n+1},
\end{align*}
with $c_i^n = c_i(\q^n)$ and $\mf{G}_i=\partial_q c_i\,\mf{N}(\q)$ the
$i\text{-th}$ row of $\mf{G}(\q)$ in \eqref{eq:full_system_dynamics.dyn}, to
approximate limit constraint forces as
\begin{align*}
    \lambda_{l,i} &= \phantom{-}\hl{\kappa}_i(\delta t + \hl{\eta}_i)(\hat{v}_{l,i} - \dot{c}^{n+1}_i)_+,\quad &\hat{v}_{l,i}=\frac{c_{l,i}-c_{i}^n}{\delta t + \hl{\eta}_i},\\
    \lambda_{u,i} &= -\hl{\kappa}_i(\delta t + \hl{\eta}_i)(\dot{c}_i^{n+1} - \hat{v}_{u,i})_+,\quad &\hat{v}_{u,i}=\frac{c_{u,i}-c_{i}^n}{\delta t + \hl{\eta}_i}.
\end{align*}

Note that the Heaviside term was neglected, allowing \emph{action-at-a-distance}
$\hl{\eta}_i\dot{c}_i^{n+1}=\beta/\pi\delta t\dot{c}_i^{n+1}$. This
$\mathcal{O}(\delta t)$ term goes to zero as the time step is decreased.

Finally, the regularized approximation contributes to \eqref{eq:sap_v} via the
cost functions
\begin{align*}
    \ell_{l,i}(\dot{c}) &= \frac{1}{2}\delta t(\delta t + \hl{\eta}_i)\hl{\kappa}_i(\dot{c}-\hat{v}_{u,i})_+^2,\\
    \ell_{u,i}(\dot{c}) &= \frac{1}{2}\delta t(\delta t + \hl{\eta}_i)\hl{\kappa}_i(\hat{v}_{l,i}-\dot{c})_+^2.
\end{align*}
Since stiffness is proportional to $\delta t^{-2}$ in \eqref{eq:near_rigid}, the
rigid approximation is tightened as the time step is reduced.

\subsection{Linear Control Laws with Effort Limits}\label{sec:icf:linear_control}

Linear effort-limited control laws can also be embedded in the convex ICF
formulation. In Sec.~\ref{sec:main:external_systems} below, we will leverage
this property to handle arbitrary external systems.

In particular, we consider control laws of the form
\begin{align}
    \mf{y} &= -\mf{C}\v + \mf{b}\nonumber\\
    \bm{\tau} &= \text{clamp}(\mf{y}; \mf{e})
    \label{eq:linear_control_with_limits}
\end{align}
where matrix $\mf{C}$ is positive diagonal, and $\mf{b}$ is a bias term. The
clamp function is defined componentwise, i.e. $\text{clamp}(\bm{\x};
\mf{e})_i=\max(-e_i, \min(e_i, x_i))$.

For such controllers, we can write the convex potential
\begin{align}
    \ell_{e,i}(v_i) &= \delta t
        \begin{cases}
            -e_i\,v_i & \text{if } y_i \geq e_i, \\
             e_i\,v_i & \text{if } y_i \leq -e_i, \\
             \frac{1}{2\,c_i} (-c_i\,v_i + b_i)^2 & \text{if } y_i\in (-e_i, e_i).
        \end{cases}\nonumber\\
    \ell_e(\v) &= \sum_i \ell_{e,i}(v_i),
	\label{eq:effort_limits_cost}
\end{align}
such that $-\nabla\ell_e$ recovers \eqref{eq:linear_control_with_limits}. ICF's
curl condition $\partial\gamma_i/\partial v_j=\partial\gamma_j/\partial v_i$ on
the control law \eqref{eq:linear_control_with_limits} leads to the requirement
that $\mf{C}$ must be positive diagonal.

\subsection{Consistency}

In total, the symplectic IMEX scheme
\eqref{eq:momentum_discrete}-\eqref{eq:q_update} is first order, with first
order approximations of compliance \eqref{eq:icf_normal_approximation} and
friction \eqref{eq:icf_friction_approximation}. Most importantly, this scheme is
\emph{consistent} with the continuous model described in
Sec.~\ref{sec:model:full}. In other words, the discrete scheme recovers exact
trajectories of the DAE \eqref{eq:full_system_dynamics} in the limit of
vanishing step size, $\delta t\to 0$. This consistency property is essential for
error-controlled integration, and allows ICF to form the core building block
upon which we develop CENIC.

\section{Convex Error-Controlled Integration}\label{sec:main}

Even advanced \mbox{$L$-stable} integrators frequently fail to advance the DAE
system~\eqref{eq:full_system_dynamics}, particularly for stiff (near-rigid)
materials and the tight friction regularization ($v_s\approx0.1$ mm) required
for robot manipulation. Numerical experiments demonstrate this in
Section~\ref{sec:experiments}. We identify three main factors underlying the
difficulties of conventional integrators. These motivate the design of CENIC,
which addresses these challenges to deliver greater robustness, accuracy, and
speed:
\begin{enumerate}
	\item \textbf{Newton convergence}. Newton--Raphson iterations in implicit
	 schemes often diverge, forcing the integrator to reject steps and discard
	 work. By building on the convexity of ICF, CENIC guarantees convergence for
	 any $\delta t$, eliminating discarded iterations.
	\item \textbf{Friction instabilities}. Friction is tightly coupled to
	tangential velocities \eqref{eq:cenic_regularized_friction}. Therefore, to
	maintain stable friction, conventional integrators must monitor velocities,
	which in turn requires small time steps to resolve impact transitions (see
	Fig.~\ref{fig:ball_drop_error}). In contrast, ICF yields well-converged
	tangential velocities and coherent friction forces at any time step,
	allowing CENIC to rely on position error as a robust accuracy metric and to
	take larger time steps (Section~\ref{sec:main:error_norm}).
	\item \textbf{Expensive geometry updates}. Standard implicit integrators—and
	some discrete
	solvers~\cite{bib:li2020ipc,lan2022affine,huang2025stiffgipc}—require costly
	geometry queries \hl{(e.g., computations of signed distances, contact
	normals, etc.)} at each Newton iteration. ICF's Taylor expansion
	\eqref{eq:icf_normal_approximation} implicitly couples forces without
	repeated queries, so CENIC needs only two geometry evaluations per step.
\end{enumerate}

Moreover, for limit and holonomic constraints, CENIC regularizes the original DAE
\eqref{eq:full_system_dynamics} using ICF in a way similar to a Singular
Perturbation Problem (SPP) \cite[\S VI]{hairer1996solving}, producing
trajectories that quickly settle onto the constraint manifold. The rapid
$\mathcal{O}(\beta \delta t)$ transient compensates for the missing freedom in
choosing consistent initial conditions in the DAE \cite[\S
VI.3]{hairer1996solving}, while error control ensures that the influence of this
transient is bounded by the desired accuracy. Furthermore, as the user-specified
accuracy tightens and time steps shrink, ICF's near-rigid regularization, which
scales as $\delta t^{-2}$ \eqref{eq:near_rigid}, yields increasingly tight
constraint enforcement.

Next we detail how CENIC leverages ICF to compute both a propagated solution and
a local error estimate. We present two strategies: a first-order scheme in
Section~\ref{sec:main:step_doubling} and a second-order scheme in
Section~\ref{sec:main:trapezoid}.

\subsection{First-Order Step-Doubling Method}
\label{sec:main:step_doubling}

To ease notation, we denote the full state update
\eqref{eq:momentum_discrete}-\eqref{eq:q_update} from state $\x^n = [\q^n;
\v^n]$ at time $t^n$ to state $\x^{n+1}=[\q^{n+1}; \v^{n+1}]$ at time
$t^{n+1}=t^n+\delta t$ as
\begin{equation}
	\x^{n+1} = \mathtt{ICF}(\x^n; \delta t).
	\label{eq:icf_update}
\end{equation}

With step-doubling, we compute the propagated solution by taking two successive
steps of size $\delta t/2$
\begin{align}
	 \x^{n+1/2} &= \mathtt{ICF}(\x^n; \delta t/2) \label{eq:half_solve_1},      \\
	 \x^{n+1} &= \mathtt{ICF}(\x^{n+1/2};\delta t/2) \label{eq:half_solve_2},
\end{align}
and compare against a full-step
\begin{align}
	\hat{\x}^{n+1} = \mathtt{ICF}(\x^n; \delta t) \label{eq:full_solve}.
\end{align}

While this requires three convex solves per step, each solve is closely related
to the others, and can be effectively warm-started
(Sec.~\ref{sec:performance:warm_start}). Moreover, quantities to build the
full-step problem are the same as those for the first half-step, and can be
reused. Thus only two geometry queries are required, since both
\eqref{eq:half_solve_1} and \eqref{eq:full_solve} use the same geometric
information at $\x^n$, while \eqref{eq:half_solve_2} requires a separate query
at $\x^{n+1/2}$.

Because ICF is first order, the resulting error estimate $e^{n+1} = \|\x^{n+1} -
\hat{\x}^{n+1}\|$ is second order \cite[\S II.4]{hairer1996solving}: overall, we
obtain a first-order integration scheme with a second-order error estimate.

\subsection{Second-Order Trapezoid Method}
\label{sec:main:trapezoid}

As above, we compute a full step $\hat{\x}^{n+1}$ from \eqref{eq:full_solve}.
From this implicit solution, we define \emph{bar} quantities that approximate
terms evaluated at $t^{n+1/2} = t^n + \delta t/2$
\begin{align*}
	 & \bar{\M} = (\M(\hat{\q}^{n+1}) + \M(\q^n))/2, \\
	 & \bar{\k} = (\k(\hat{\q}^{n+1}) + \k(\q^n))/2, \\
	 & \bar{\N} = (\N(\hat{\q}^{n+1}) + \N(\q^n))/2,
\end{align*}
and write our second-order scheme as
\begin{subequations}\label{eq:second_order}
\begin{align}
    &\bar{\M}(\v^{n+1} - \v^n) + \delta t \bar{\k} = \overline{\J^T\bgamma}, \\
    & \q^{n+1} = \q^n + \frac{\delta t}{2} \bar{\N}(\v^{n+1} + \v^n).
	\label{eq:second_order_q_update}
\end{align}
\end{subequations}

Without contact ($\overline{\J^T\bgamma} = 0$), \eqref{eq:second_order}
corresponds to a second-order explicit trapezoid rule. With contact, the key is
to obtain a second-order approximation of the contact contribution
$\overline{\J^T\bgamma}$. For stability, we seek an implicit approximation in
$\v^{n+1}$. To that end, we propose a scheme based on the implicit trapezoid
method
\begin{eqnarray*}
    \overline{\J^T\bgamma} &\defeq&
    \frac{1}{2}\left({\J^n}^T\bgamma^n+{\J^{n+1}}^T\bgamma^{n+1}\right)\\
    &\approx&\frac{1}{2}\left(\mf{j}^n + \hat{\J}^T\bgamma^{n+1}\right),
\end{eqnarray*}
where $\mf{j}^n={\J^n}^T\bgamma^n$ are the generalized impulses at the
\hl{beginning of the step} and \emph{hat} quantities are next time step
estimates from $\hat{\x}^{n+1}$. We treat the Jacobian explicitly, $\hat{\J} =
\J(\hat{\q}^{n+1})$, reusing values from the initial full step. We treat the
impulses implicitly, setting up an incremental potential to compute
$\bgamma^{n+1}=\bgamma(\v^{n+1};\hat{\x}^{n+1}, \delta t)$. 

Unlike \eqref{eq:constraint_cost_gradient}, this incremental potential uses the
state at $t^{n+1}$ rather than at $t^n$. Therefore we must introduce
approximations implicit in $\v^{n+1}$ that can use $\hat{\x}^{n+1}$. We define
\begin{equation*}
    \hat{f}_{e,i}^n=f_{e,i}(\hat{\q}^{n+1})+\delta t\,k_i\,\hat{v}_{n,i}
\end{equation*}
to be used in \eqref{eq:icf_normal_approximation}, where $\hat{v}_{n, i}$ is the
normal component of $\hat{\vf{v}}_{c,i}=\hat{\J}_i\hat{\v}^{n+1}$. That is,
$\hat{f}_{e,i}^n$ is a first-order estimation of the elastic normal force 
\emph{stepped backwards} from $\hat{\q}^{n+1}$.

Comparing our second-order scheme with \eqref{eq:momentum_discrete}, we notice
that it corresponds to an ICF problem \eqref{eq:icf_cost} with
\begin{align*}
    &\mf{A} = \mf{\bar{\M}},\\
    &\mf{r} = \bar{\M}\v^n-\delta t \bar{\k} + \frac{1}{2}\mf{j}^n,\\
    &\J = \frac{1}{2}\hat{\J},
\end{align*}
and an incremental approximation for $\bgamma^{n+1}$ from $\hat{f}_{e,i}^n$. We
solve this ICF problem for $\v^{n+1}$ and update configurations
with \eqref{eq:second_order_q_update} to achieve second-order accuracy.

Finally, we compare the second-order $\x^{n+1}$ with the first-order
$\hat{\x}^{n+1}$ to obtain a second-order error estimate. In total, this
second-order trapezoid method requires only two convex ICF solves and two
geometry queries at each step: one at $\x^n$ to build the first-order update and
one at $\hat{\x}^{n+1}$ to build the second-order update. As we will show in
Sec.~\ref{sec:experiments:error_estimation}, this second-order method improves
energy conservation, but at the expense of losing desirable stability
properties (in particular, the implicit trapezoid rule is not $L$-stable).

\subsection{External Systems}\label{sec:main:external_systems}

Stiff external systems, such as a high-gain PID controller, can force
error-controlled integration to take many small time steps, slowing simulation
speeds. A more complex example is the vacuum gripper in Section
\ref{sec:experiments:external_systems}, where vacuum forces increase rapidly as
function of distance to the suction caps, thus leading to stiff dynamics and
short time scales.

To handle such cases, which are common in robotics, this section develops an
implicit scheme for integrating external dynamical systems tightly coupled to
the multibody dynamics. The resulting scheme can be applied to arbitrary
external systems, enabling the stable treatment of complex controllers and
user-defined force elements.

The key idea is to use first-order approximations of the external system
dynamics \eqref{eq:external_system_dynamics} to express actuation inputs
$\btau^{n+1}$ as an implicit function of $\v^{n+1}$ in the linear form
\eqref{eq:linear_control_with_limits}, compatible with ICF.

\begin{remark}
	Similarly, we could use second-order approximations of the external system
	dynamics compatible with our second-order ICF. However, as we will show in
	Section~\ref{sec:experiments}, the step-doubling strategy is much more
	effective in practice and thus we present only the first-order approximation
	here.
\end{remark}

First, we use \eqref{eq:q_update} to write a first-order approximation about
$\q^n$ that eliminates configurations from \eqref{eq:external_system_dynamics}
\begin{align*}
	&\dot{\z}(\z, \v) \approx \mf{h}(\z, \tilde{\x}(\v)) = \tilde{\mf{h}}(\z, \v),
\end{align*}
with $\tilde{\x}(\v)=[\q^n+\delta t\N^n \v, \v]$. Then we linearize the
external system dynamics around $\z^n$ and $\v^{n}$ to obtain
\begin{align}
	\dot{\z}(\z, \v) & \approx \partial_z\tilde{\mf{h}}^n \z + \partial_v\tilde{\mf{h}}^n \v + \mf{b}_h^n,\nonumber\\
	\mf{b}_h^n & = \tilde{\mf{h}}(\z^n, \v^n) - \partial_z\tilde{\mf{h}}^n \z^n - \partial_v\tilde{\mf{h}}^n \v^n.
	\label{eq:external_dynamics_linearization}
\end{align}
As in a Rosenbrock integration scheme \cite[\S IV.7]{hairer1996solving}, we
integrate the linearized dynamics implicitly as
\begin{equation}
	\z^{n+1} = \z^n + \delta t \left( \partial_z\tilde{\mf{h}}^n \z^{n+1} + \partial_v\tilde{\mf{h}}^n \v^{n+1} + \mf{b}_h^n \right).
	\label{eq:discrete_external_dynamics}
\end{equation}

Note that for linear external systems, such as PID controllers, the
linearization \eqref{eq:external_dynamics_linearization} is exact and the
Rosenbrock approximation reduces to implicit Euler.

\begin{remark}
	In cases where the external system dynamics are sufficiently simple, we can
	also consider an explicit Euler scheme, which results from replacing
	\eqref{eq:external_dynamics_linearization} with $\dot{\z} =
	\tilde{\mf{h}}(\z^n, \v^n)$. This eliminates the cost of computing
	$\partial_z\tilde{\mf{h}}^n$ and $\partial_v\tilde{\mf{h}}^n$, but may force
	error control to take smaller steps if the external system dynamics
	\eqref{eq:ext_sys_dynamics} are stiff.
\end{remark}

We solve $\z^{n+1}$ as a function of $\v^{n+1}$ from
\eqref{eq:discrete_external_dynamics}
\begin{equation}
	\z^{n+1} = \mf{Z}\,\v^{n+1} + \mf{b},
	\label{eq:z_from_x}
\end{equation}
with
\begin{align*}
	\mf{Z} &= \delta t\left( \mathbf{I} - \delta t \partial_z\tilde{\mf{h}}^n \right)^{-1}\,\partial_v\tilde{\mf{h}}^n,\\
	\mf{b} &= \left( \mathbf{I} - \delta t \partial_z\tilde{\mf{h}}^n \right)^{-1}\left( \z^n + \delta t \mf{b}_h^n \right).
\end{align*}

Similarly, we now use \eqref{eq:q_update} and \eqref{eq:z_from_x} to eliminate
$\q$ and $\z$ from the actuation \eqref{eq:ext_sys_output} to produce
\begin{equation*}
	\btau^{n+1} \approx \mf{g}(\mf{Z}\,\v^{n+1} + \mf{b}, \tilde{\x}(\v^{n+1})) = \tilde{\mf{g}}(\v^{n+1}),
\end{equation*}
and linearize around $\v^n$ to obtain the approximation
\begin{equation*}
	\btau^{n+1} = \partial_v\tilde{\mf{g}}^n(\v^{n+1}-\v^n) + \tilde{\mf{g}}^n.
\end{equation*}
To obtain a linear control law compatible with
\eqref{eq:linear_control_with_limits}, we make one final approximation
\begin{equation}
	\btau^{n+1} = -\mf{C}\v^{n+1} + \mf{d},
	\label{eq:u_from_v}
\end{equation}
with 
\begin{align*}
	&\mf{C} = \max(0, \mathrm{diag}\!\left(-\partial_v\tilde{\mf{g}}^n\right)),
	&\mf{d} = \tilde{\mf{g}}(\v^n) + \mf{C}\v^n.
\end{align*}

For stabilizing diagonal linear external systems, including PID controllers,
$\mf{C}=-\partial_v\tilde{\mf{g}}^n$ and we recover an implicit Euler scheme.
Moreover, we can also handle (highly non-linear) effort limits using the
convex potential \eqref{eq:effort_limits_cost}. In practice, most well-designed
control systems are stabilizing. Thus, for non-linear control strategies such as
\emph{inverse dynamics PID} or \emph{task space control}, $\mf{C}$ is a good
approximation of $-\partial_v\tilde{\mf{g}}^n$ that greatly improves simulation
stability. In the rare case of stiff but non-stabilizing external forces where
$\mf{C}=\mf{0}$, \eqref{eq:u_from_v} resolves to the explicit approximation
$\btau^{n+1}\approx\btau^n$, and error control will shrink the time step
accordingly to resolve the dynamics at the requested accuracy.

We compute $\partial_v\tilde{\mf{g}}^n$ with forward differences, though more
sophisticated alternatives like automatic differentiation are also possible.
Finally, with $\v^{n+1}$ computed from the convex ICF, we advance the external
state $\z^{n+1}$ using \eqref{eq:z_from_x}.

\begin{figure}
	\centering
	\includegraphics[width=\linewidth]{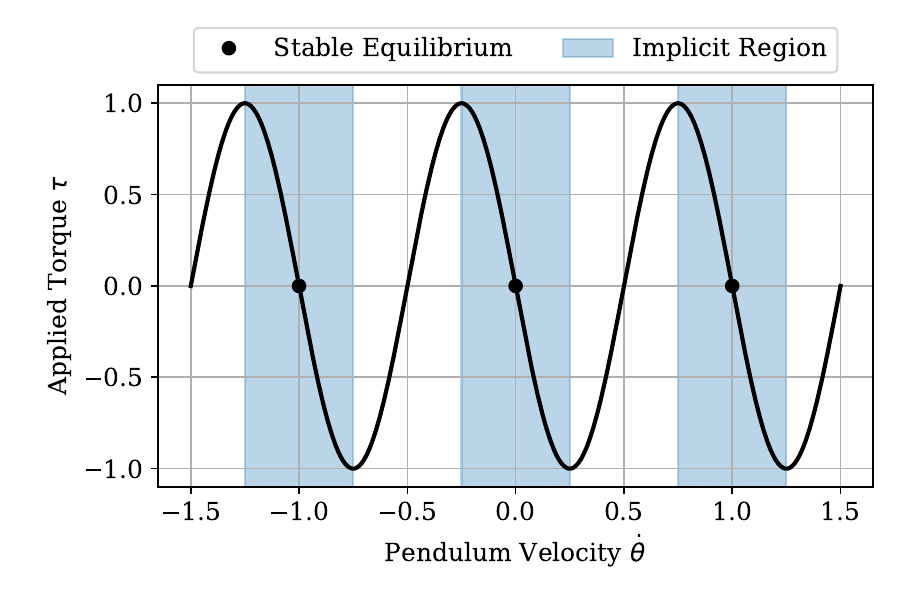}
	\caption{\hl{A simple nonlinear controller \eqref{eq:sine_controller}
	induces a series of stable and unstable equilibria in a torque-controlled
	pendulum. CENIC's linearization strategy integrates the controller
	implicitly in the stable (shaded, negative slope, negative feedback)
	regions. To maintain convexity, we fall back to an explicit scheme in the
	unstable regions. Enforcing convexity everywhere while performing implicit
	integration whenever possible enables faster and more reliable simulation of
	complex nonlinear controllers, see
	Sec.~\ref{sec:experiments:external_systems}. }}
	\label{fig:sine_controller}
\end{figure}

\hl{
\begin{example}
	Consider a simple torque-controlled pendulum with state $\x=[\theta;
	\dot{\theta}]$, driven by an external controller
	\begin{equation}\label{eq:sine_controller}
	    \tau = -\sin(2 \pi \dot{\theta}).
	\end{equation}
	This nonlinear controller induces stable equilibria around $\dot{\theta} =
	0, 1, 2, \ldots$. As illustrated in Fig.~\ref{fig:sine_controller}, these
	equilibria are surrounded by regions of {\em negative feedback}. It is
	precisely in these negative feedback regions where the linearization matrix
	$\mf{C}$ is non-zero, providing implicit integration. In the unstable
	(positive feedback) regions, $\mf{C}=\mf{0}$ and the scheme falls back to an
	explicit treatment of the nonlinear controller.
\end{example}
}

\hl{
	Our diagonalization and clipping strategy for $\mf{C}$ is a simple way to
	ensure that the resulting ICF problem is convex. More sophisticated
	alternatives, such as a full SVD decomposition and projection, may offer
	tighter approximations at greater computational cost. Thoroughly
	characterizing such tradeoffs is an important direction for future work.
	Nonetheless, experiments in Section~\ref{sec:experiments:external_systems}
	demonstrate that even a simple diagonal approximation yields significant
	stability improvements for complex nonlinear external systems.
}

\hl{
Importantly, external system linearization does not change the consistency of
the overall integration scheme. 

\begin{proposition}
	The convex ICF problem with external system linearization
	\eqref{eq:u_from_v} is consistent and accurate to first-order, i.e., the
	local truncation error scales as $\mathcal{O}(\delta t^2)$.
\end{proposition}

\begin{proof}[Proof (sketch)]
	In the absence of external systems, we know that ICF is consistent and
	first-order accurate \cite{castro2024irrotational}. The key is to show that
	linearization error does not change this first-order accuracy. To
	illustrate this, consider the simplified system
	\begin{equation*}
		\dot{\x} = \mf{g}(\x)
	\end{equation*}
	and the integration scheme
	\begin{equation*}
		\x^{n+1} = \x^n + \delta t (\mf{g}(\x^n) + \mf{C} (\x^{n+1} - \x^n)),
	\end{equation*}
	where $\mf{C}$ is any finite matrix. This is a simplified version of
	\eqref{eq:u_from_v}, where we treat the external system as the only source
	of dynamics. Rearranging, we have
	\begin{equation*}
		\x^{n+1} - \x^n = \delta t (\mf{I} - \delta t \mf{C})^{-1} \mf{g}(\x^n),
	\end{equation*}
	where
	\begin{equation*}
		(\mf{I} - \delta t \mf{C})^{-1} = \mf{I} + \delta t \mf{C} + \mathcal{O}(\delta t^2).
	\end{equation*}
	Plugging back into the integration scheme, we have
	\begin{align*}
		\x^{n+1} &= \x^n + \delta t \mf{g}(\x^n) + \mathcal{O}(\delta t^2) \\
		&= \x^n + \int_{t^n}^{t^{n+1}} \mf{g}(\x(t)) dt + \mathcal{O}(\delta t^2),
	\end{align*}
	where the second line follows from the consistency of the forward Euler
	method.

	Similar reasoning applies to our case with full multibody dynamics with
	contact: external system linearization does not change the order or
	consistency of the integration scheme, regardless of any approximations that
	go into computing $\mf{C}$.
\end{proof}
}

\subsection{Static and Dynamic Friction}\label{sec:main:friction}

Mathematically, it is impossible for discrete formulations to disambiguate
between static and dynamic friction. Ad-hoc extensions to PGS-style solvers from
computer graphics~\cite{bib:muller2020} are sometimes applied to robotics
\cite{physx, bib:raisim}, but without mathematical grounding these methods lack
convergence guarantees and can produce hard-to-quantify artifacts. 

In contrast, continuous-time formulations offer a rigorous framework for
capturing both static and dynamic friction through regularized friction. In
particular, we use the model \eqref{eq:cenic_regularized_friction}, which
captures both regimes. However, the negative slope between $\mu_s$ and $\mu_d$
breaks the convexity of ICF.

To incorporate the non-convex friction model
\eqref{eq:cenic_regularized_friction} into our convex formulation, we \emph{lag}
the regularized term $\mu(s)$ as
\begin{equation*}
	\vf{f}_t^{n+1} = -\mu(\Vert\vf{v}_t^n\Vert/v_s)\frac{\vf{v}_t^{n+1}}{\sqrt{\Vert\vf{v}_t^{n+1}\Vert+v_s^2}}f_n^n.
\end{equation*}

This approximation is convex, converges to the continuous model
\eqref{eq:cenic_regularized_friction} as time step goes to zero (i.e. it is
\emph{consistent}), and maintains a tight approximation in stiction (when
$\Vert\vf{v}_t^{n+1}\Vert < v_s$). CENIC keeps the local truncation error
introduced by this approximation within the user-requested accuracy by
modulating the time step as needed.

\subsection{The Error Norm}\label{sec:main:error_norm}

When components of the state $\x$ differ in scale or units, the error estimate
can become dominated by certain terms while others are negligible. This is
common in multibody systems, where $\x = [\q; \v]$ includes both positions
and velocities.

\begin{figure}
	\centering
	\includegraphics[width=\linewidth]{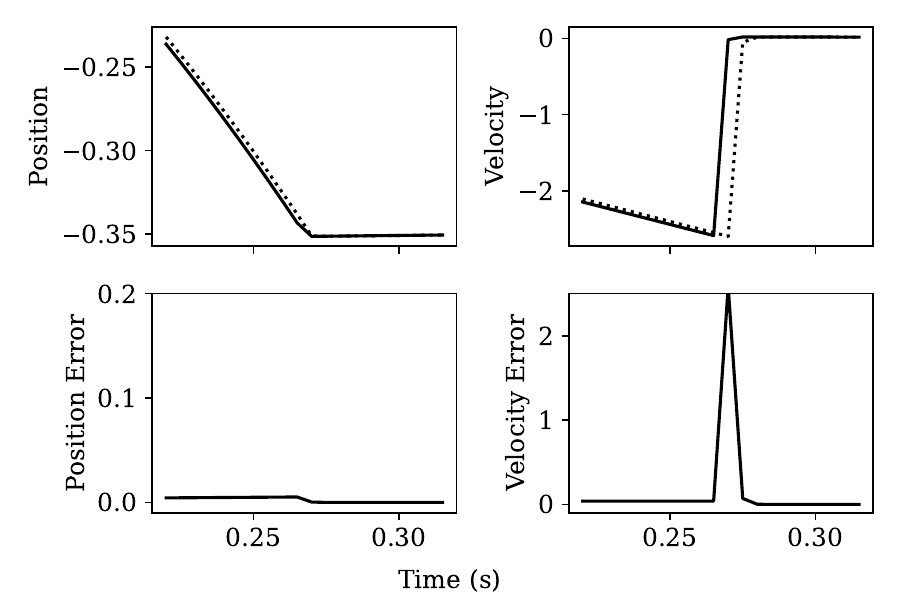}
	\caption{ Vertical position and velocity of a ball dropped on a table,
		showing differences between solutions computed with a 1~ms (solid line) and a
		5~ms (dashed line) time step. Velocity error spikes, but position error
		does not. }
	\label{fig:ball_drop_error}
\end{figure}

As an example, consider a ball dropped onto a table.
Figure~\ref{fig:ball_drop_error} shows vertical position, velocity, and
corresponding errors for two simulations with different (fixed) time steps.
While the trajectories are similar overall, the velocity error exhibits a sharp
spike at impact due to the impulsive nature of contact. Even with compliant
contact, such rapid velocity changes dominate the error estimate, forcing the
integrator to take small steps to capture fast time scales.

To address this issue, we measure error using a weighted position-only
$L^{\infty}$ norm
\begin{equation*}
e^{n+1} = \Vert\mf{S}(\q^{n+1} - \hat{\q}^{n+1})\Vert_\infty,
\end{equation*}
where the diagonal scaling matrix $\mf{S}$ maps each component to a
dimensionless unit. This makes $\varepsilon_\text{acc}$ in
\eqref{eq:step_size_udpate} an acceptable fraction of unit error, or
equivalently, the desired \emph{digits of accuracy} in the solution
\cite{sherman2011simbody} (0.1 roughly corresponding to 10\%). $\mf{S}$ can be
estimated from knowledge of coordinate types or it can be specified by expert
users.

By excluding impulsive velocity jumps, this norm focuses the error estimate on
positions—capturing artifacts such as rattling and passthrough evident in $\q$.
As shown in Section~\ref{sec:experiments:work_precision_plots}, this choice
reduces simulation times by over an order of magnitude in contact-rich problems. 

\subsection{Method Summary}\label{sec:main:summary}

Algorithm~\ref{alg:cenic} summarizes CENIC. Users provide initial conditions,
desired accuracy $\varepsilon_\text{acc}$, and a maximum time step size $\delta
t_\text{max}$.

\begin{algorithm}
	\caption{CENIC}
	\label{alg:cenic}
	\DontPrintSemicolon
	\KwIn{$\x^0$, $\z^0$, $\varepsilon_\text{acc}$, $\delta t_\text{max}$, $T$}

	$t \gets 0$

	$n \gets 0$

	$\delta t \gets k_\text{Init}\,\delta t_\text{max}$

	\While{$t \le T$} {
		Linearize external systems: \hfill (Sec.~\ref{sec:main:external_systems})

		\quad $\mf{Z}, \mf{b}, \mf{C}, \mf{d} = \mathtt{LinearizeExtSys}(\x^n, \z^n)$

		Perform convex optimization: \hfill (Sec.~\ref{sec:main:step_doubling}-\ref{sec:main:trapezoid})

		\quad $\x^{n+1}, \hat{\x}^{n+1} = \mathtt{DoStep}(\x^n; \delta t, \mf{C}, \mf{d})$

		Advance external systems state:
	
		\quad $\z^{n+1} = \mf{Z} \v^{n+1} + \mf{b}$

		Estimate the error: \hfill (Sec.~\ref{sec:main:error_norm})

		\quad $e^{n+1} = \|\mf{S}(\q^{n+1} - \hat{\q}^{n+1})\|_\infty$

		\If{$e^{n+1} \leq \varepsilon_\text{acc}$} {

			Accept step and propagate solution:

			\quad $t \gets t + \delta t$			

			\quad $\x^{n}\gets\x^{n+1}$
	
			\quad $\z^{n}\gets\z^{n+1}$

			\quad $n \gets n + 1$
		}
		Update the time step:

		\quad $\delta t \gets \mathtt{AdjustStepSize}(\delta t; e^{n+1}, \varepsilon_\text{acc})$
	}
\end{algorithm}

At each step, CENIC linearizes external system dynamics
\eqref{eq:external_system_dynamics} around the current state ($\x^n, \z^n$).
This is used by \texttt{DoStep()} to compute the solution $\x^{n+1}$ and error
reference $\hat{\x}^{n+1}$. \texttt{DoStep()} uses either the first-order
step-doubling method (Section~\ref{sec:main:step_doubling}) or the second-order
trapezoid method (Section~\ref{sec:main:trapezoid}). If the error estimate
(Section~\ref{sec:main:error_norm}) is below the desired accuracy, the
solution is propagated and time is advanced forward. 

In practice, \texttt{AdjustStepSize()} uses standard step-selection
logic\cite[\S 4.8]{hairer1996solving}, implementing a dead-band function to
introduce hysteresis and prevent repeated step size switches. More precisely
\begin{align*}
	&\delta\hat{t} = \mathtt{Deadband}(k_\text{Safe}\delta t(\varepsilon_\text{acc}/e^{n+1})^{1/p}, \delta t),\\
	&\delta t \gets \min(\delta\hat{t}, k_\text{MaxGrow}\,\delta t, \delta t_\text{max}),
\end{align*}
where
\begin{equation*}
	\mathtt{Deadband}(\delta t_\text{new}, \delta t) = 
	\begin{cases}
		\delta t & \text{if } k_\text{Low}\,\delta t < \delta t < k_\text{High}\,\delta t,\\
		\delta t_\text{new} & \text{Otherwise }.
	\end{cases}
\end{equation*}

A mature implementation must also include other safety checks for corner cases
such as divide-by-zero or not-a-number (NAN) errors. We use $k_\text{Init}=0.1$,
$k_\text{Safe}=0.9$, $k_\text{Low}=0.9$, $k_\text{High}=1.2$ and
$k_\text{MaxGrow}=5.0$.

\begin{table*}
	\centering
	\hl{
	\begin{tabular}{r|c|c|c}
		& \textbf{Discrete Solvers} \cite{bib:bullet,todorov2012mujoco,physx,bib:raisim,Genesis} & \textbf{Error-Controlled Implicit Integrators} \cite{hairer1996solving} & \textbf{CENIC} \\
		\hline
		\textbf{Simulate Stiff Frictional Contact} & Fast & Very Slow & Fast \\
		\textbf{Guaranteed Accuracy} & No & Yes & Yes \\
		\textbf{Convex} & Sometimes & No & Yes \\
		\textbf{Geometry Queries} & One per step & One per iteration (many per step) & Two per step \\
		\textbf{Reduce Timestep} & Never & For accuracy and convergence failure & For accuracy only \\
		\textbf{Implicit External Systems} & No & Yes & Yes \\
		\textbf{Static \& Dynamic Friction} & No & Yes & Yes \\
	\end{tabular}
	}
	\caption{\hl{Comparison of key characteristics of different simulation
	strategies. CENIC brings together the best of classical
	\hll{error-controlled} integration and discrete solvers often used
	for robotics simulation.} 
	\label{tab:method_comparison}}
\end{table*}
\section{Performance Optimizations}\label{sec:performance}

In this section, we introduce several performance optimizations enabled by
error-controlled integration. At each step, CENIC solves a series of convex
problems of the form
\begin{subequations}\label{eq:final_optimization}
\begin{align}
    &\v^{n+1} = \min_{\v} \ell(\v), \\
    &\ell(\v) = \frac{1}{2}\v^T\A\v -\mf{r}^T\v + \ell_c(\v).
\end{align}
\end{subequations}

We solve each convex problem iteratively with Newton's method. At the
$i\text{-th}$ iteration, the Newton update is
\begin{subequations}\label{eq:newton_solve}
\begin{align}
    & \H_i \p_i = -\g_i  \\
    & \v_{i+1} = \v_i + \alpha_i \p_i,
\end{align}
\end{subequations}
where $\H_i = \nabla^2 \ell(\v_i)$ is the Hessian, $\g_i = \nabla \ell(\v_i)$
the gradient, $\p_i$ the search direction, and $\alpha_i$ is computed with exact
linesearch \cite{castro2022unconstrained}. The optimizations described below aim
to solve this sequence of problems more efficiently.

\subsection{Warm Starts}\label{sec:performance:warm_start}

Primal formulations like \eqref{eq:final_optimization} are easily warm started
from prior solutions. For the step-doubling scheme
(Section~\ref{sec:main:step_doubling}), we warm start the full step to
$\hat{\v}^{n+1}$ with $\v^n$, the first half-step to $\v^{n+1/2}$ with $(\v^n +
\hat{\v}^{n+1})/2$, and the second half-step to $\v^{n+1}$ with
$\hat{\v}^{n+1}$. For the trapezoid scheme (Section~\ref{sec:main:trapezoid}),
we warm start the first order solution $\hat{\v}^{n+1}$ with $\v^n$, and the
second order solution $\v^{n+1}$ with $\hat{\v}^{n+1}$.

\subsection{Adaptive Convergence Criteria}\label{sec:performance:convergence}

Robotics simulations typically require tight convergence to ensure stable
contacts \cite{castro2022unconstrained,todorov2012mujoco}. With error control,
however, Newton iterations \eqref{eq:newton_solve} only need to converge to a
tolerance that is negligible compared to the requested accuracy
\cite{hairer1996solving}.

CENIC uses two convergence criteria. The first checks the optimality condition
\begin{equation}\label{eq:gradient_tolerance}
    \|\mf{D} \g\| \leq \epsilon_{tol} \max(1, \|\mf{D} \mf{r}\|),
\end{equation}
where $\mf{D} = \mathrm{diag}(\M)^{-1/2}$ scales all variables to have the same
units \cite{castro2022unconstrained} and $\epsilon_{tol}$ is a dimensionless
tolerance. This criterion allows for early exit even before any Newton
iterations take place, in case the initial guess is a solution within the required
tolerance.

The second criterion evaluates the convergence of Newton iterations to the true
solution $\v^*$. Following \cite[p. 121]{hairer1996solving}, the true error
$\Vert\v^* -\v_{i+1}\Vert$ is estimated from the norm of $\Delta \v_{i} =
\v_{i+1} - \v_i$. This leads to the criterion
\begin{equation}\label{eq:velocity_change_tolerance}
    \eta_i \|\mf{D}^{-1} \Delta \v_i\| \leq \epsilon_{tol} \max(1, \|\mf{D} \mf{r}\|),
\end{equation}
where $\mf{D}^{-1}$ scales velocities to the same units as
\eqref{eq:gradient_tolerance}, and
\begin{align*}
    \Theta_i &= \|\Delta \v_{i+1}\| / \|\Delta \v_{i}\|,\\
    \eta_i   &= \Theta_i / (1 - \Theta_i).
\end{align*}

In fixed-step mode, we use a tight tolerance $\epsilon_{tol} = 10^{-8}$
suitable for stable discrete-time simulation. In error controlled mode, we
instead set 
\begin{equation}
    \epsilon_{tol} = \max(\kappa \cdot \varepsilon_\text{acc}, 10^{-8}),
\end{equation}
for some $\kappa \in [0, 1]$. We found $\kappa = 0.001$ to provide a reasonable
balance between performance and stable contact resolution across all of our test
cases.

\subsection{Hessian Reuse}\label{sec:performance:hessian_reuse}

Since consecutive convex solves differ only slightly, we reuse factorizations in
analogy to Jacobian reuse in implicit Runge--Kutta methods \cite[\S
IV.8]{hairer1996solving}. Unlike implicit Runge--Kutta, however, CENIC does not
require monitoring Newton's method for divergence, as ICF's convexity
\cite{castro2024irrotational} guarantees convergence for any positive definite
Hessian approximation, albeit possibly slowly for poor approximations. 

\begin{figure}
    \centering
    \includegraphics[width=\linewidth]{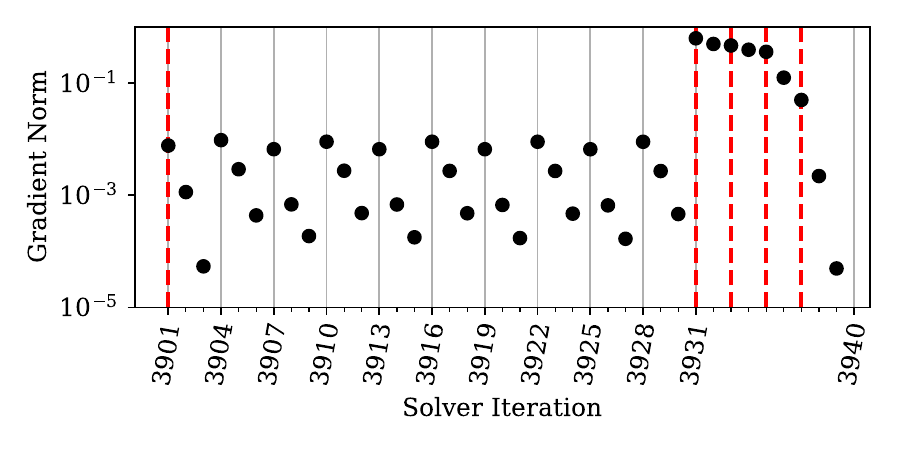}
    \caption{Solver convergence during hard clutter simulation, illustrating the
    impact of Hessian reuse. Red dashed lines indicate a recomputed Hessian,
    while light grey lines separate time steps.
    \eqref{eq:hessian_reuse_criterion} allows us to reuse the same Hessian when
    possible (first ten steps). When necessary, a few fresh factorizations
    enable rapid Newton-type convergence (final step).
    }
    \label{fig:convergence}
\end{figure}

Instead, we estimate the convergence error after a desired maximum number of
iterations $N$. If this satisfies criterion
\eqref{eq:velocity_change_tolerance}, we can keep reusing the current Hessian
approximation. Otherwise, the Hessian is recomputed. The actual number of
iterations can be higher, $N$ is simply a \emph{desired} maximum.

Assuming linear convergence rate $\Theta_i$ (conservative for Newton), the true
error $\Vert\v^* -\v_{i+M}\Vert$ after $M$ additional iterations is
$\Theta^{M}_i/(1 - \Theta_i)\|\Delta \v_i\|$. Thus we recompute the Hessian if
\begin{equation}\label{eq:hessian_reuse_criterion}
    \frac{\Theta^{N - i}_i}{1 - \Theta_i}\|\Delta \v_i\| \geq \epsilon_{tol} \max(1, \|\mf{D} \mf{r}\|),
\end{equation}
where $N = 10$ in our implementation. Using this criterion, we reuse Hessians
not only across iterations, but also across convex solves and time steps.

Figure~\ref{fig:convergence} shows the convergence history of a test simulation
from Sec.~\ref{sec:experiments}, illustrating the effectiveness of the approach.
Reusing the same factorization across time steps yields fast but only linear
convergence; when this reuse becomes limiting, the method detects it, recomputes
the Hessian, and recovers quadratic convergence. 

\subsection{Cubic Linesearch Initialization}\label{sec:performance:linesearch}

Our exact linesearch uses Newton-Raphson with bisection fallback \cite[\S
9.4]{press2007numerical}. Although the required first and second derivatives can
be evaluated in linear time \cite[\S IV.C]{castro2022unconstrained}, linesearch
remains a significant cost (Sec.~\ref{sec:experiments:performance_profiling}).

A better initial guess can reduce this cost. In
\cite{castro2022unconstrained,castro2024irrotational}, the initial guess is from
a quadratic approximation of $\ell_\alpha(\alpha) := \ell(\v + \alpha p)$ using
$\ell_\alpha(0)$, $\ell_\alpha'(0)$, and $\ell_\alpha''(0)$. In this work, we
instead use a cubic approximation, constructed from $\ell_\alpha(0)$,
$\ell_\alpha'(0)$, $\ell_\alpha(\alpha_\text{max})$, and
$\ell'_\alpha(\alpha_\text{max})$. Typically $\alpha_\text{max}=1$, though
values as large as $\alpha_\text{max}=1.5$ also work well.

\section{Experimental Results}\label{sec:experiments}

\begin{figure*}
    \centering
    \begin{subfigure}{0.19\linewidth}
        \centering
        \includegraphics[width=\linewidth]{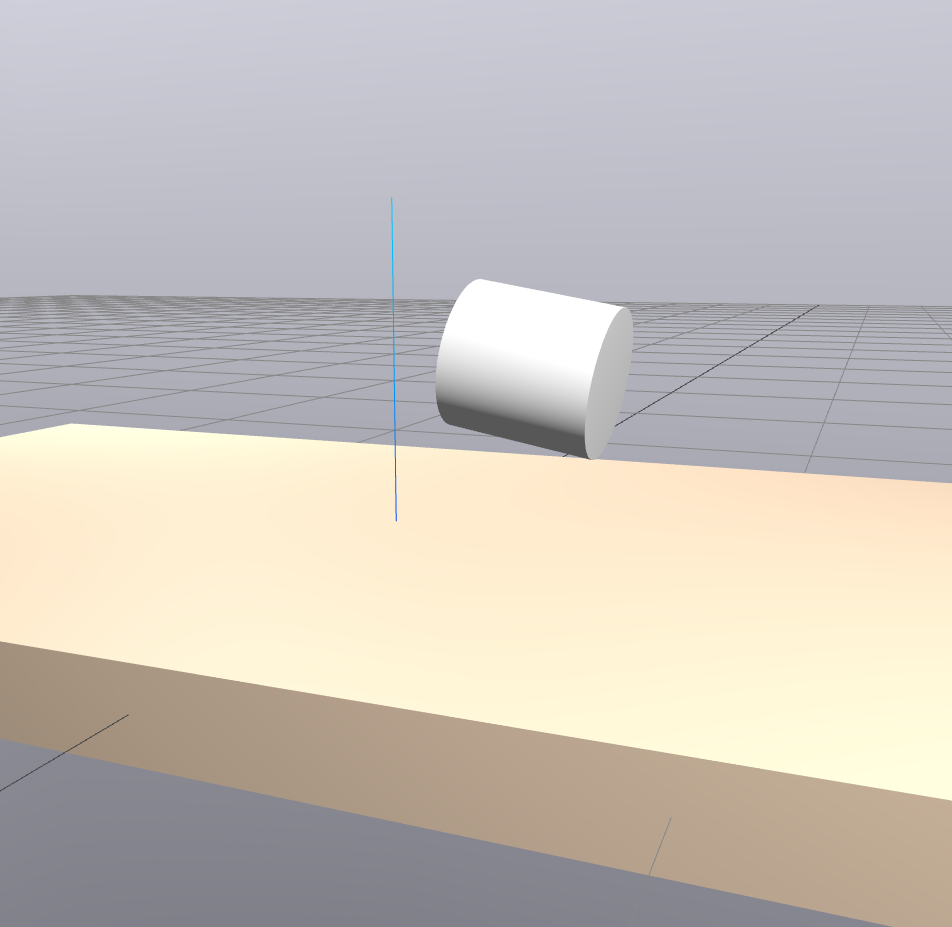}
        \caption{\textbf{Cylinder} falls on a table.}
        \label{fig:examples:cylinder}
    \end{subfigure}
    \begin{subfigure}{0.19\linewidth}
        \centering
        \includegraphics[width=\linewidth]{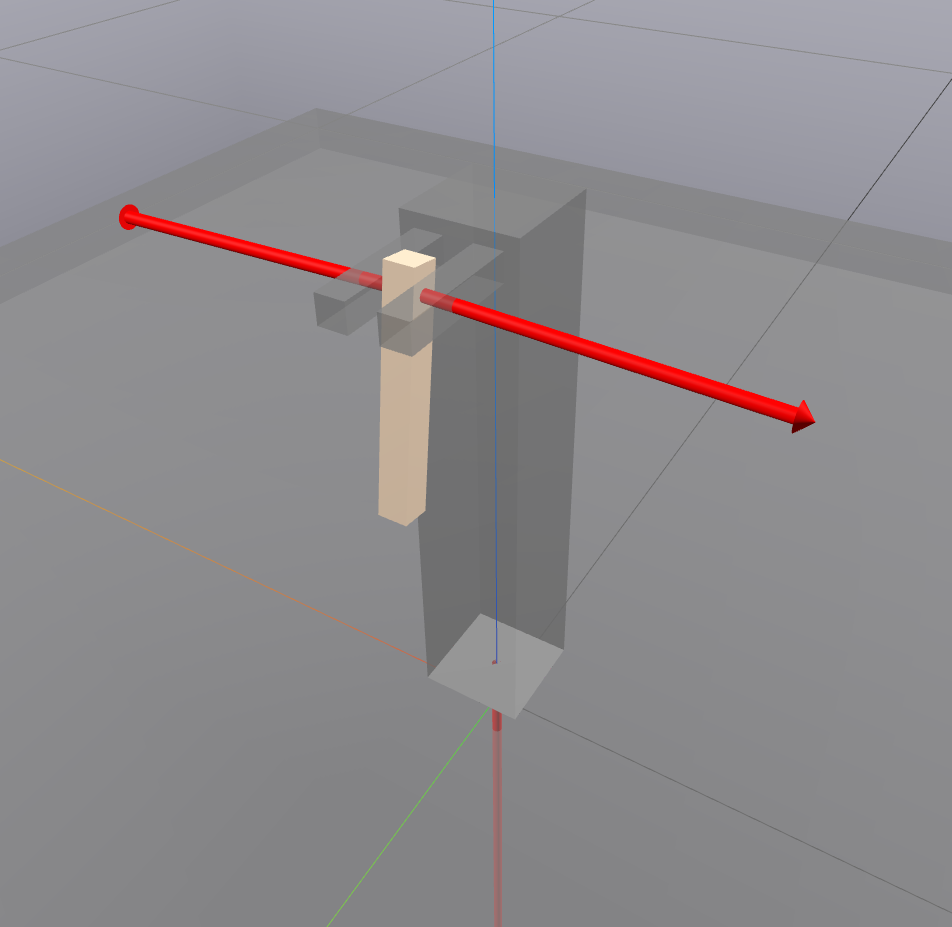}
        \caption{\textbf{Static gripper} holds a peg.}
        \label{fig:examples:gripper}
    \end{subfigure}
    \begin{subfigure}{0.19\linewidth}
        \centering
        \includegraphics[width=\linewidth]{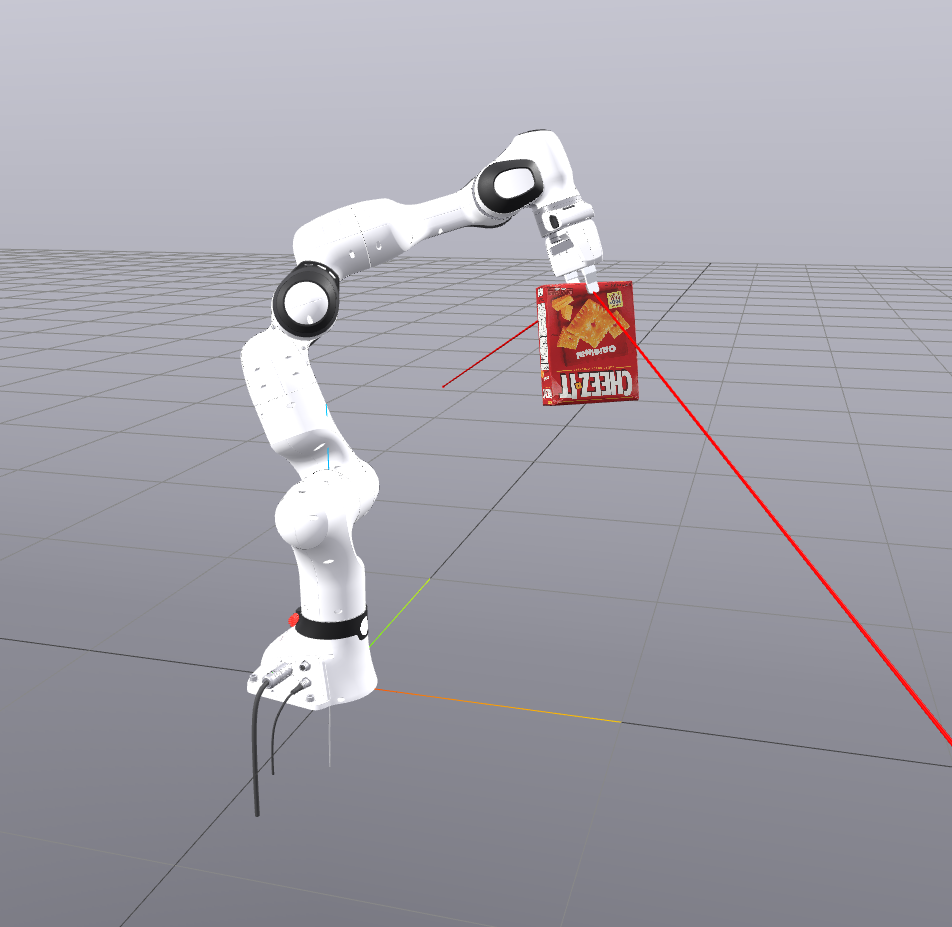}
        \caption{\textbf{Franka Panda} moves a box.}
        \label{fig:examples:franka}
    \end{subfigure}
    \begin{subfigure}{0.19\linewidth}
        \centering
        \includegraphics[width=\linewidth]{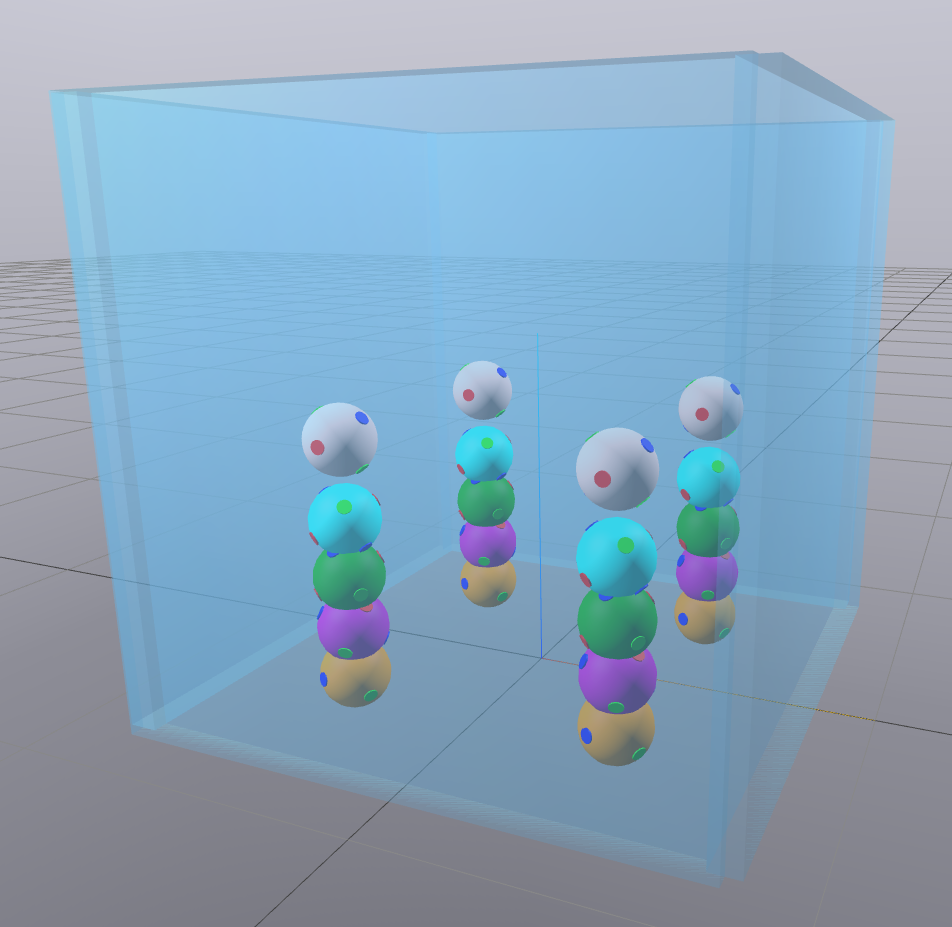}
        \caption{\textbf{Soft clutter} falls in a box.}
        \label{fig:examples:soft_clutter}
    \end{subfigure}
    \begin{subfigure}{0.19\linewidth}
        \centering
        \includegraphics[width=\linewidth]{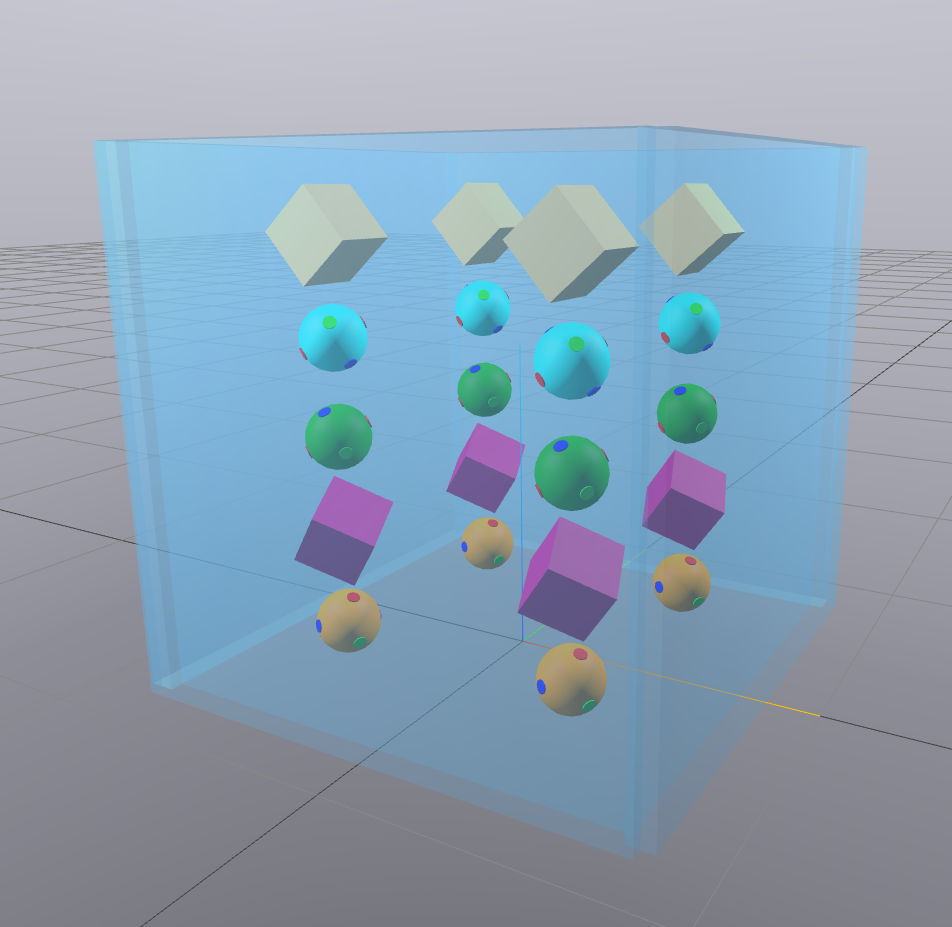}
        \caption{\textbf{Hard clutter} falls in a box.}
        \label{fig:examples:hard_clutter}
    \end{subfigure}
    \caption{Simulation test cases, ranging from least to most complex. These
        test cases highlight different challenges in robotics simulation: sharp
        corners (\subref{fig:examples:cylinder}), stable force closure
        (\subref{fig:examples:gripper}), stiff nonlinear controllers
        (\subref{fig:examples:franka}), many degrees-of-freedom
        (\subref{fig:examples:soft_clutter}), and complex contact interactions
        (\subref{fig:examples:hard_clutter}).}
    \label{fig:examples}
\end{figure*}

This section presents simulation results characterizing the performance of
CENIC. Throughout this section, we refer to five test cases shown in
Fig.~\ref{fig:examples}. These scenarios highlight different challenges in
robotics simulation \hl{and span a wide range of problem sizes, as summarized
in Table~\ref{tab:example_scales}.}

\begin{table}
    \centering
    \hl{
    \begin{tabular}{|c|c|c|}
        \hline
        \textbf{Name} & \textbf{Degrees of Freedom} & \textbf{Max. Num.
        Contacts} \\
        \hline
        cylinder & 6 & 1 \\
        gripper & 7 & 3 \\
        franka & 16 & 2 \\
        soft clutter & 120 & 31 \\
        hard clutter & 120 & 77 \\
        \hline
    \end{tabular}
    }
    \caption{\hl{Scale summary (degrees of freedom and maximum number of
    simultaneous contacts) for the test cases shown in
    Fig.~\ref{fig:examples}.}}
    \label{tab:example_scales}
\end{table}

First, a \textbf{cylinder} falls onto a table
(Fig.~\ref{fig:examples:cylinder}). While this example is simple, the cylinder's
sharp corners produce forces that vary abruptly with changes in state. The
cylinder has radius 0.1~m, length 0.2~m, and mass 6.3~kg.

\begin{figure}
    \centering
    \includegraphics[width=\linewidth]{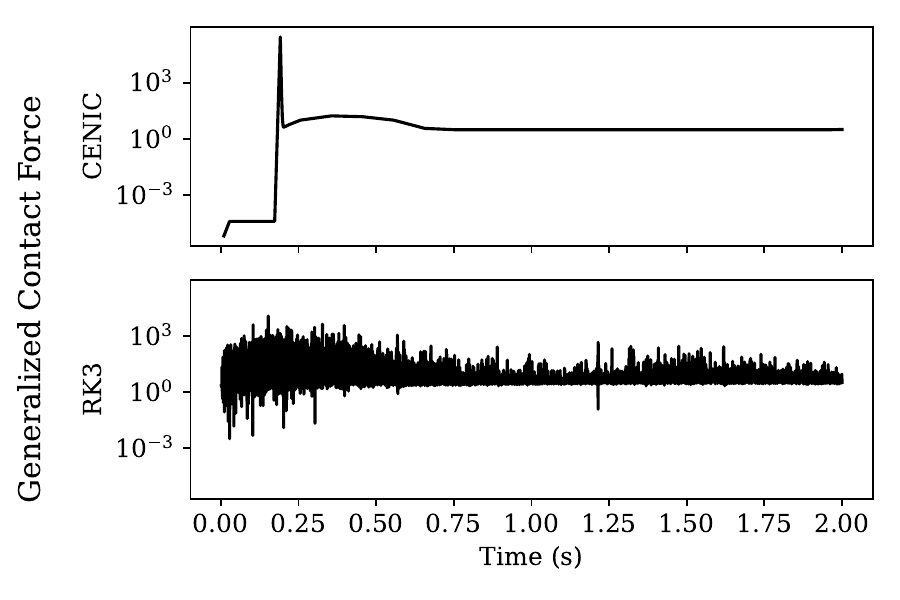}
    \caption{Contact force magnitude for the static gripper example at accuracy
    $\varepsilon_\text{acc} = 10^{-3}$. CENIC (top) produces smooth, stable
    contact forces, while RK3 (bottom) produces noisy, unstable results.
    Deceivingly, positions might look visually accurate with RK3, until later in
    the simulation when RK3 mistakenly drops the object.}
    \label{fig:gripper_forces}
\end{figure}

The next case isolates grasping complexity to its simplest form---a peg wedged
between two stationary fingers of a \textbf{static gripper}
(Fig.~\ref{fig:examples:gripper}). The gripper slides along a prismatic joint
and impacts the ground; the peg should remain wedged through this impulsive
event. Despite the apparent simplicity of this scenario, many methods exhibit
unstable friction oscillations, as shown in Fig.~\ref{fig:gripper_forces}. The
peg is a $1\times1\times8\text{~cm}$ block. Each finger is a
$1\times1\times8\text{~cm}$ block spaced such that the gap between them is
$0.01\text{~mm}$ narrower than the peg's width. The fingers are rigidly attached
to a $4\times4\times20\text{~cm}$ post. All bodies have the density of water.

The \textbf{Franka} example (Fig.~\ref{fig:examples:franka}) reflects a typical
robotics workflow. A high-gain inverse-dynamics controller drives a Franka Emika
Panda arm between two setpoints while the gripper grasps a
Cheez-It\textsuperscript{\textregistered} box. Beyond testing force closure
between the gripper and box, this case also requires simulating a stiff
user-defined controller. As discussed in
Section~\ref{sec:main:external_systems}, the high controller gains introduce
short time scales that challenge numerical stability.

The \textbf{soft clutter} and \textbf{hard clutter} examples test scalability by
dropping 20 objects into a bin. The soft clutter case uses only spheres, a
contact stiffness of $10^3~\text{N/m}$, and a large friction regularization
($v_s = 1~\text{cm/s}$). The hard clutter case mixes spheres and cubes, with
stiffer contact parameters approximating rigid Coulomb friction: contact
stiffness $10^5~\text{N/m}$ and stiction tolerance $0.1~\text{mm/s}$.

The first three examples (cylinder, gripper, franka) use hydroelastic contact
\cite{elandt2019pressure,masterjohn2022velocity}, while both clutter examples
use point contact. All tests were completed on the same workstation with an
Intel Xeon w7-3565X CPU. 

\subsection{Comparing First and Second-Order Methods}\label{sec:experiments:error_estimation}

\begin{figure}
    \centering
    \includegraphics[width=\linewidth]{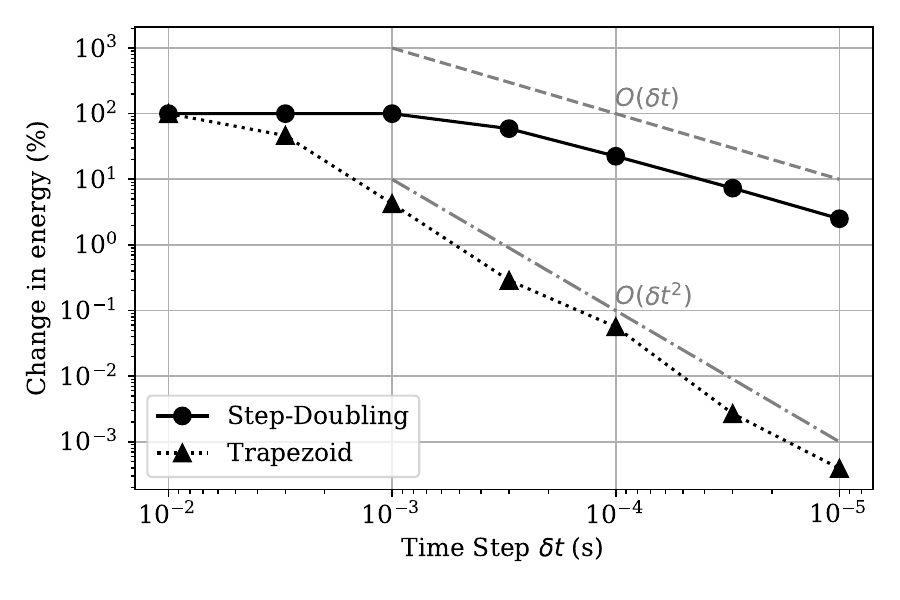}
    \caption{Energy conservation error (percent of energy lost after 10 seconds)
        for a 0.1 kg bouncing ball with zero dissipation. The ball is relatively
        soft (contact stiffness $10^3$ N/m) and bounces 11 times in the 10
        second simulation. Potential energy is defined such that total energy is
        zero when the ball is at rest on the ground. The trapezoid method
        (Sec.~\ref{sec:main:trapezoid}) shows second-order convergence, while
        step-doubling (Sec.~\ref{sec:main:step_doubling}) is first order.}
        
    \label{fig:energy_conservation}
\end{figure}

\begin{figure}
    \centering
    \includegraphics[width=\linewidth]{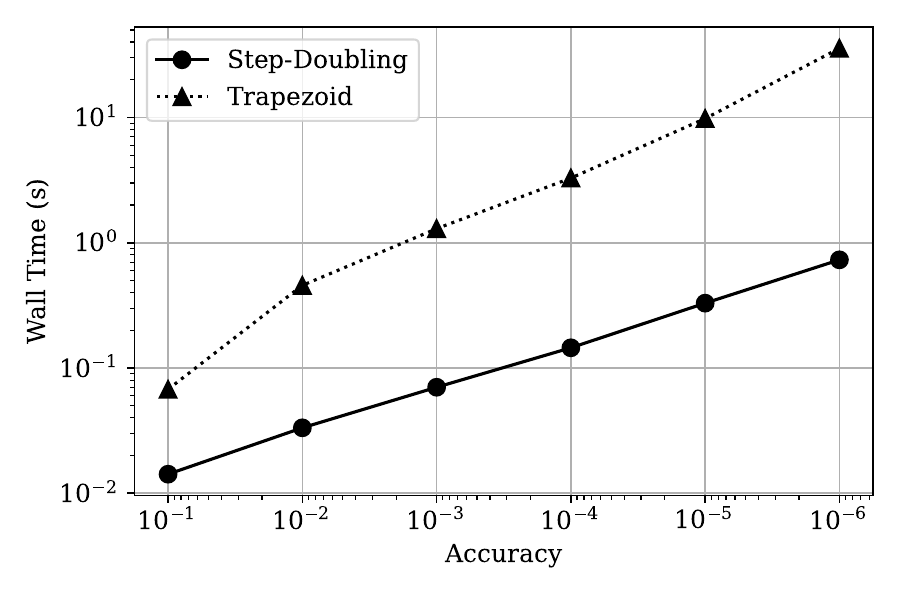}
    \caption{Work-precision plots for hard clutter (lower is better). Despite
    requiring only two convex solves per step versus three for step-doubling,
    the trapezoid method's lack of stability hinders performance.}
    \label{fig:error_wp}
\end{figure}

As shown in Fig.~\ref{fig:energy_conservation}, the second-order method improves
energy conservation through contact, where a 1 kg ball with zero
dissipation is dropped from an initial height of 1 m. The energy conservation
error decays quadratically with $\delta t$ under the trapezoid method, but only
linearly with step-doubling.

But this benefit is offset by a loss of stability, which forces the trapezoid
method to take much smaller time steps to control stability rather than
accuracy. As shown in Fig.~\ref{fig:error_wp}, the second-order trapezoid method
is consistently 5-20 times slower than step-doubling on the hard clutter
example, despite requiring fewer convex solves per time step. For this reason,
and because strict energy conservation is rarely a critical requirement for
robotics simulation, we \textbf{prefer the first-order step-doubling} approach
and use it for the remainder of this paper.

\hl{In the absence of contact and other constraints, CENIC with step-doubling
reduces to a first-order semi-implicit Euler method, which does provide some
energy-preserving symplectic properties \cite{hairer1996solving}. Further
research is required to extend such properties to the constrained/contact-rich
case.}

\subsection{Comparison with Other Error-Controlled Integrators}\label{sec:experiments:work_precision_plots}

\begin{figure*}
    \centering
    \includegraphics[width=0.9\linewidth]{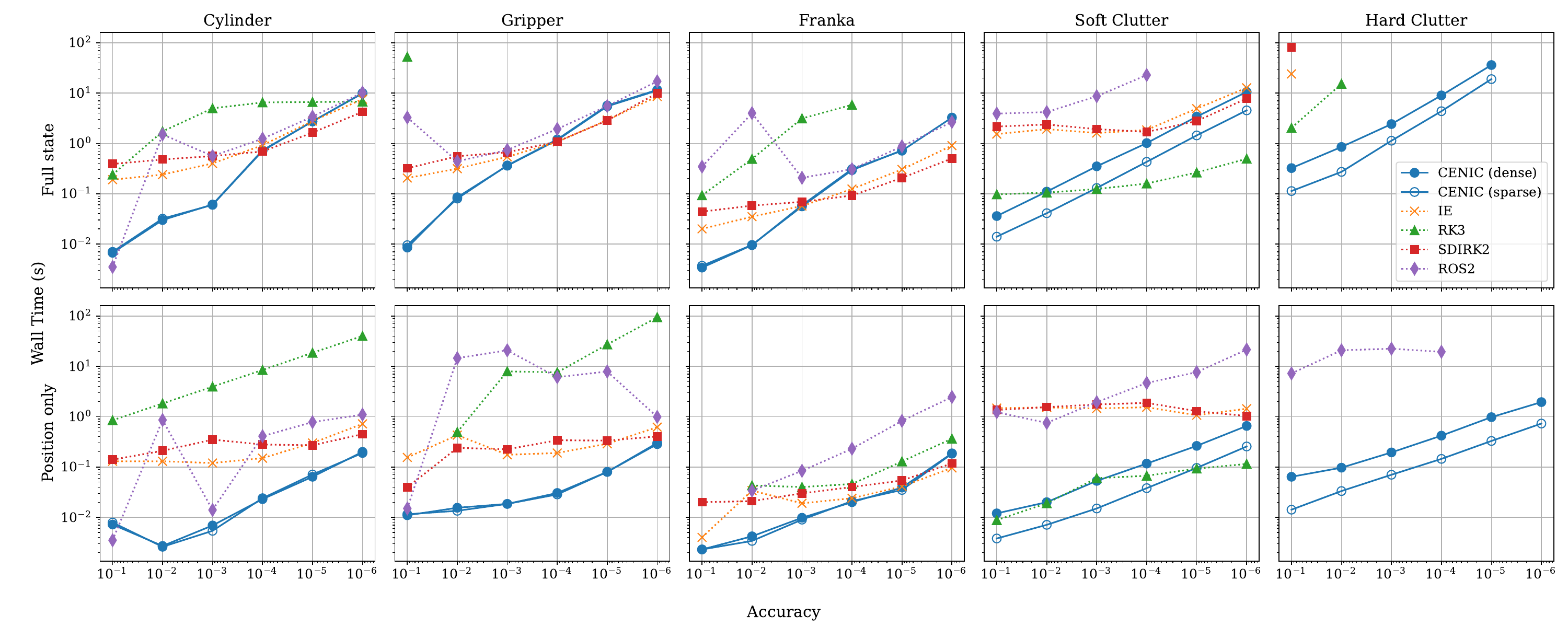}
    \caption{Work-precision plots for the test cases in Fig.~\ref{fig:examples},
    using error control on the full state $\x$ (top) and on positions $\q$ only
    (bottom). Lower is better. Wall times are normalized to one second of
    simulation time. Missing data points indicate solver failure or timeout
    after 100 seconds (real-time rate $< 1\%$). CENIC performs particularly well
    with many contacts and at loose accuracies, where convergence failures force
    other implicit schemes to take small steps. }
    \label{fig:examples_wp}
\end{figure*}

We compare CENIC with several state-of-the-art error-control baselines. Our
first baseline is implicit Euler (\textbf{IE}) with step-doubling for error
estimation,  a first-order \mbox{$L$-stable} scheme. We use a mature
implementation with Jacobian reuse and accuracy-driven stopping criteria
\cite[\S IV.8]{hairer1996solving}. We also compare with a third-order
Runge--Kutta scheme (\textbf{RK3}). Unlike IE, RK3 is an explicit scheme, and is
not \mbox{$L$-stable}.

For higher order \mbox{$L$-stable} baselines, we compare with the
\textbf{SDIRK2} method \cite{kennedy2016diagonally} and the two-stage Rosenbrock
method \textbf{ROS2} \cite{blom2016comparison}, both second-order. Rosenbrock
methods are essentially implicit Runge--Kutta methods that only take one Newton
iteration per step \cite{hairer1996solving}. Rosenbrock methods are known to be
particularly effective on stiff problems at loose accuracies. Our implementation
is based on \cite{zhang2014fatode}.

Figure~\ref{fig:examples_wp} shows work-precision plots comparing these four
baselines with CENIC. All implementations are in C++ using Drake \cite{drake}.
\hl{This helps isolate any differences in performance to the underlying
algorithm, rather than implementation details or quality.} Since our baseline
integrators use only dense algebra, for a fair comparison we report CENIC
results with both sparse and dense Cholesky factorizations. In addition to
comparing integration schemes across examples, Fig.~\ref{fig:examples_wp} also
illustrates the impact of computing an error estimate over the full state (top)
and over positions only (bottom), as discussed in
Section~\ref{sec:main:error_norm}.

While the baseline integration schemes produce acceptable results for the
simpler examples (e.g., cylinder, gripper), only CENIC is able to simulate the
hard clutter scenario faster than real time\hl{, i.e., the ten second simulation
takes less than ten seconds of wall time}. Additionally, considering
position error enables a consistent order-of-magnitude speedup for CENIC, but
not for the other baselines. This is likely due to the tightly converged
frictional contact solution produced by ICF in a single step, while other
integrators must take many small time steps to maintain stability and resolve
stick-slip transitions. Moreover, ICF requires a single geometric query per
convex solve, while standard integrators need to update geometric quantities at
every Newton iteration.

\subsection{Comparison with Discrete-Time Robotics Simulators}\label{sec:experiments:comparison}

\begin{figure*}
    \centering
    \includegraphics[width=0.85\linewidth]{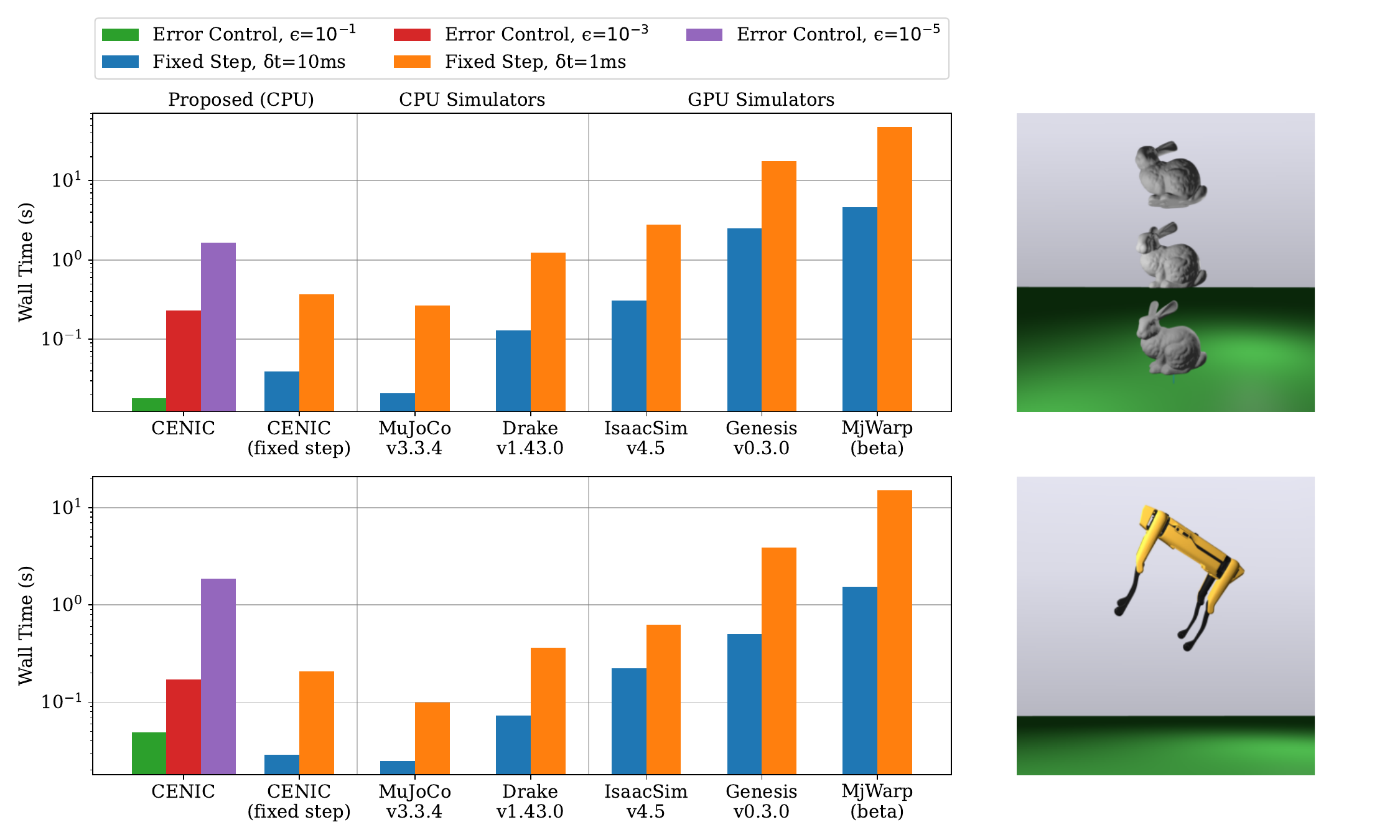}
    \caption{Simulator speed comparison\hl{: wall times to complete a 10-second
    (top) and 3-second (bottom) simulation}. Three mesh bunnies (top) and an
    unactuated quadruped robot (bottom) are dropped on the ground. CENIC is
    competitive with state-of-the-art robotics simulators in both fixed-step and
    error-controlled mode. Note that these benchmarks measure speed only, not
    simulation quality. Also note that the GPU simulators are optimized for
    simulating many parallel scenes, but these tests use only a single scene.}
    \label{fig:simulator_comparison}
\end{figure*}

We compare simulation speeds with two test cases shown in
Fig.~\ref{fig:simulator_comparison}: three large mesh bunnies (height
$1\text{~m}$, mass $200\text{~kg}$) and an unactuated Boston Dynamics Spot
model, both dropped on the ground. The Spot model is from the MuJoCo Menagerie
\cite{menagerie2022github}, while the bunny model is from the Stanford 3D
Scanning Repository \cite{stanford3drepository}. The bunnies are simulated for
10 seconds (at which point they cease rolling), while the quadruped is simulated
for 3 seconds. 

We compare \textsc{CENIC} against discrete solvers in
MuJoCo~\cite{todorov2012mujoco}, IsaacSim~\cite{physx}, Genesis~\cite{Genesis},
and Drake~\cite{drake}. Except for the time step, we use default simulator
parameters. Figure~\ref{fig:simulator_comparison} shows wall times for all
simulators using 1~ms and 10~ms fixed time steps, as well as for CENIC with
accuracies $\varepsilon_\text{acc} \in [10^{-1}, 10^{-3}, 10^{-5}]$. Remarkably,
CENIC achieves comparable simulations speeds, even though it is tasked to
compute a solution to desired accuracy.

\textbf{This test evaluates speed only, not quality.} All simulators produced
visually plausible results, though their differing parameterizations and
stiffness choices lead to distinct trade-offs between speed and physical
realism. For example, when simulating the Spot model with a 1~ms time step,
MuJoCo required 7613 solver iterations but allowed up to 3.1~cm of ground
penetration, whereas CENIC (in fixed-step mode) required 9056 iterations but
produced only 6.4~mm of penetration. A full analysis of these trade-offs and
their implications for sim-to-real transfer is an important area for future
work, but outside the scope of this paper.

\subsection{Novel Capabilities}\label{sec:experiments:unique_capabilities}

We highlight several capabilities of CENIC that are difficult, or even
mathematically impossible, to incorporate into existing discrete-time
simulators.

\subsubsection{\textbf{Stiff External Systems}}
\label{sec:experiments:external_systems}

As discussed in Section~\ref{sec:main:external_systems}, \textsc{CENIC}
integrates external systems implicitly. This allows for modular design: users
can define arbitrarily complex systems, including those with internal state,
without modifying or recompiling the solver or any multibody component. In
contrast, popular robotics simulators offer limited support for modeling
controllers (often only PD control), while complex user-defined subsystems must
be handled explicitly, leading to unstable coupling with the multibody dynamics.

\begin{figure}
    \centering
    \includegraphics[width=0.9\linewidth]{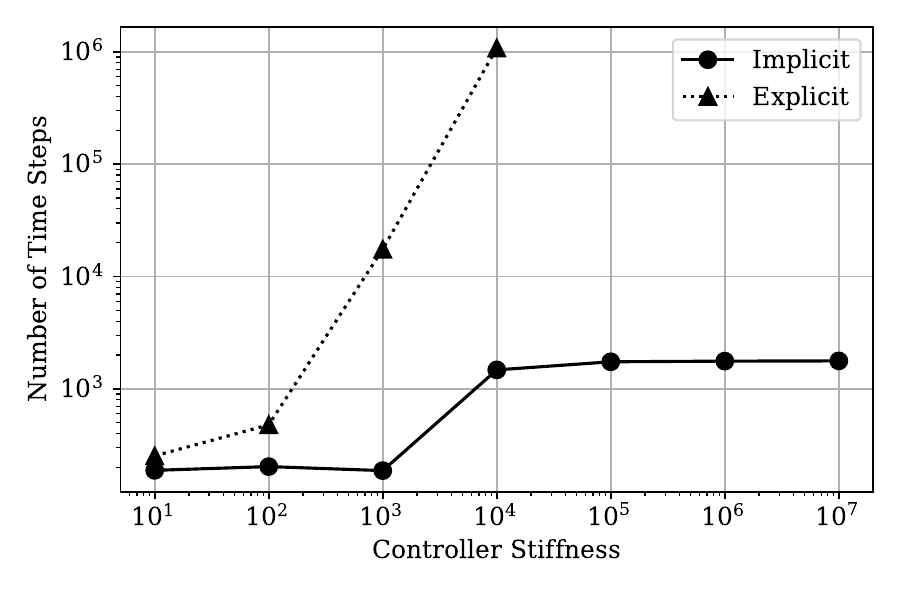}
    \caption{Number of time steps required to simulate the franka task for 10
    seconds with different controller stiffnesses ($ = K_p = K_d$) and accuracy
    $\varepsilon_\text{acc} = 0.01$. With a standard explicit treatment, step size must
    decrease to keep stability under control. In contrast, our proposed implicit
    scheme (Sec.~\ref{sec:main:external_systems}) is stable even with extremely
    stiff controllers.}
    \label{fig:implicit_external_sys}
\end{figure}

As a first example, the Franka arm in Fig.~\ref{fig:examples:franka} tracks
desired positions $\q_d$ and velocities $\v_d$ using an inverse dynamics
controller
\begin{align*}
    &\btau = \M(\q)\dot{\v}_d + \k(\q, \v),\\
    &\dot{\v}_d = -K_p(\q - \q_d) - K_d(\v-\v_d),
\end{align*}
where $K_p$ and $K_d$ are proportional and derivative gains, respectively.
Figure \ref{fig:implicit_external_sys} contrasts an explicit treatment of this
nonlinear external system (as done in most simulators) with our proposed
diagonally-implicit approach. Our scheme allows CENIC to handle extremely high
gains without the need to take more time steps---step size is solely determined
by the requested accuracy, not stability.

\begin{figure}
    \centering
    \begin{subfigure}{0.3\linewidth}
        \centering
        \includegraphics[width=\linewidth]{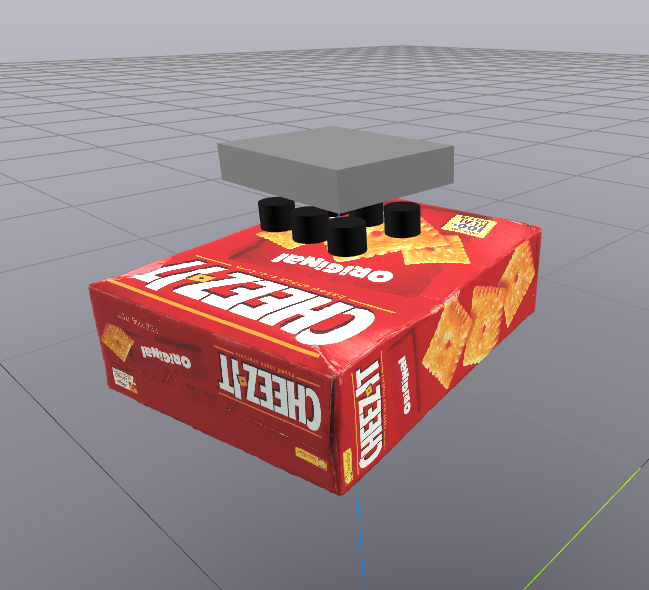}
        \caption{Initial condition}
        \label{fig:suction:init}
    \end{subfigure}
    \begin{subfigure}{0.3\linewidth}
        \centering
        \includegraphics[width=\linewidth]{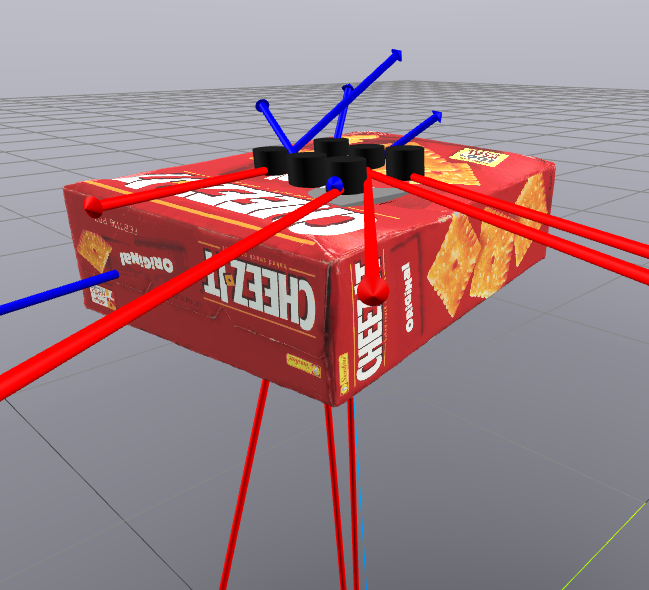}
        \caption{Discrete time \cite{castro2024irrotational}}
        \label{fig:suction:sap}
    \end{subfigure}
    \begin{subfigure}{0.3\linewidth}
        \centering
        \includegraphics[width=\linewidth]{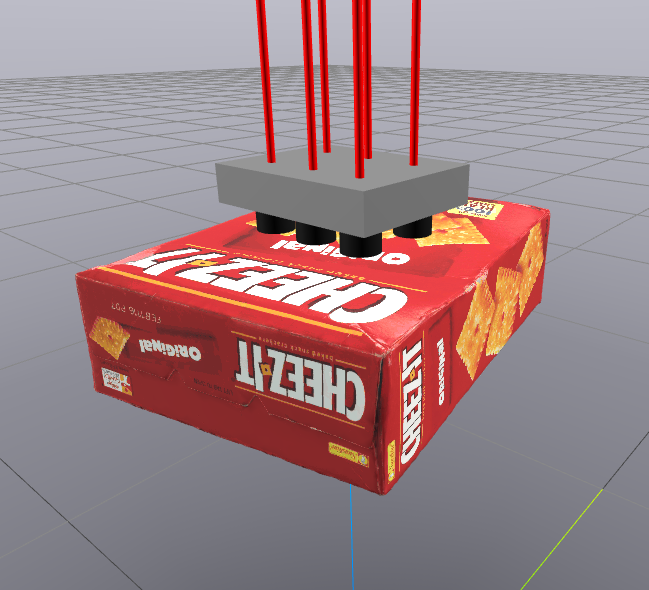}
        \caption{CENIC (ours)}
        \label{fig:suction:convex}
    \end{subfigure}
    \caption{Suction gripper after 0.5s of simulation. For discrete
    time-stepping, we use $\delta t = 0.01$ s. For CENIC, we request accuracy
    $\varepsilon_\text{acc} = 10^{-3}$. Large forces pulling the box into the
    gripper lead to massive passthrough artifacts in the discrete-time
    simulation. CENIC resolves the dynamics correctly with a combination of
    error control (Section~\ref{sec:main:step_doubling}) and implicit treatment of
    the gripper force model (Section~\ref{sec:main:external_systems}).}
    \label{fig:suction}
\end{figure}

This modular capability extends beyond controller modeling.
Figure~\ref{fig:suction} illustrates a vacuum gripper composed of an array of
bellows, each modeled as a small-mass cylinder attached to a spring-loaded
prismatic joint. Suction from each bellow is \hl{a simple force pulling on
nearby objects,
\begin{equation}
    f^{suction} = \max\left(0,f^{max}\left(1 - \frac{\phi(\q)}{r}\right)\right),
\end{equation}
where $f^{max} = 50$ N is the maximum suction force, $r = 4$ cm is the suction
radius, and $\phi(\q)$ is the signed distance to the nearest point on an
object.} Bellows are pressed against the pad as objects are caught by the
vacuum. \hl{While this is an exceedingly simple suction model, it captures the
key difficulties in simulating external systems:} the combination of light
masses and strong suction forces makes this system particularly challenging for
standard discrete-time approaches.

With discrete time-stepping, large forces treated explicitly lead to severe
passthrough (Fig.~\ref{fig:suction:sap}). In contrast, CENIC advances the
solution without artifacts (Fig.~\ref{fig:suction:convex}).

\subsubsection{\textbf{Static and Dynamic Friction}}

CENIC provides, for the first time, a mathematically rigorous treatment of
distinct static and dynamic friction coefficients that is compatible with the
high speed and scalability requirements of robotics simulation
(Section~\ref{sec:main:friction}).

\begin{figure}
    \centering
    \includegraphics[width=\linewidth]{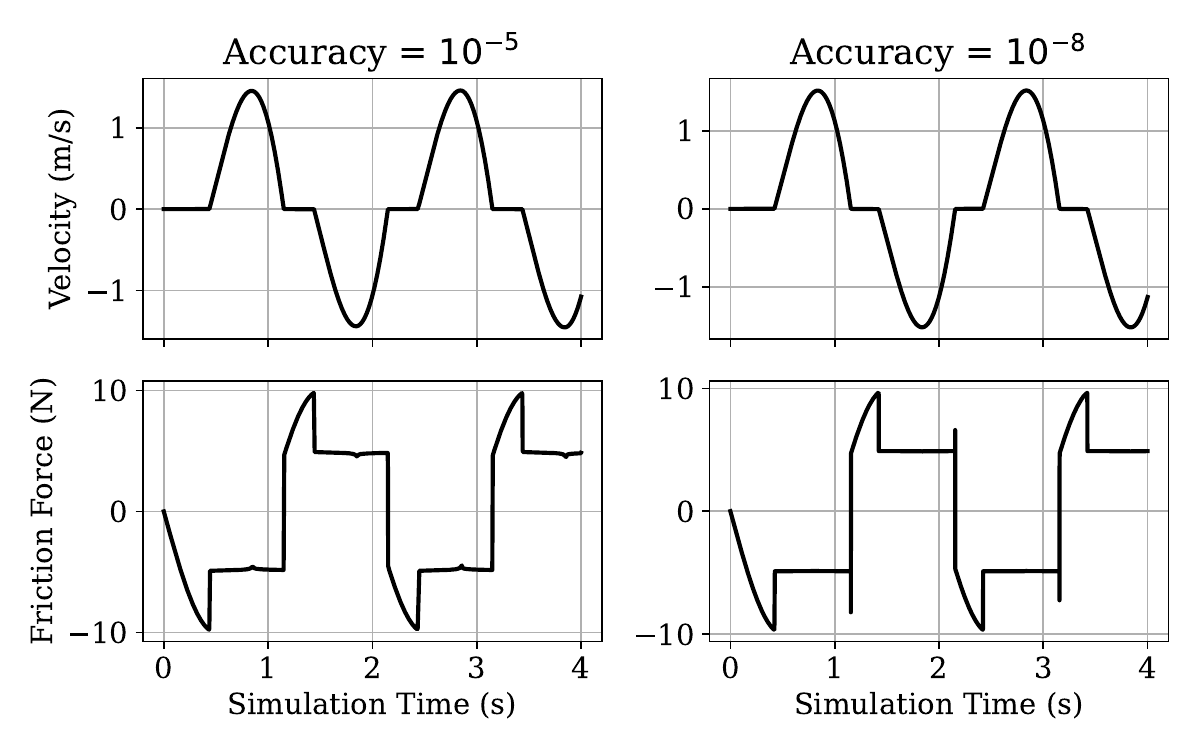}
    \caption{Tangential velocities and friction forces for a 1 kg box on an
    oscillating conveyor belt, demonstrating transitions between static ($\mu_s
    = 1.0$) and dynamic ($\mu_d = 0.5$) friction. An accuracy of
    $10^{-5}$ (left) shows the box ``catching'' in stiction at low velocities. A
    tight accuracy ($10^{-8}$, right) is needed to capture the impulsive
    transition from sliding to stiction. Existing convex time-stepping methods
    are unable to capture any of these effects, as they only support a single
    coefficient of friction.}
    \label{fig:static_dynamic_friction}
\end{figure}

To demonstrate this, we simulate a 1~kg block sitting on an oscillating conveyor
belt. Friction coefficients between belt and box are $\mu_s = 1.0$ and $\mu_d =
0.5$. Figure~\ref{fig:static_dynamic_friction} plots tangential friction force
and velocity relative to the conveyor belt. The box transitions between stiction
and sliding, with contact force jumps due to distinct friction coefficients. As
accuracy is tightened, impulsive transitions from sliding back to stiction
become evident.

\subsection{Hardware Validation}\label{sec:experiments:hardware_validation}

To evaluate CENIC in a practical manipulation scenario, we consider a
real-to-sim task involving numerous challenging objects (Fig.~\ref{fig:hero}),
inspired by the most demanding benchmarks used to assess large behavior model
(LBM) performance in simulation~\cite{barreiros2025careful}. In this
\textbf{dish rack} scenario, a teleoperated robot drops several thin objects
into a plastic bin, then dumps the contents of the bin on a wire dish rack. We
then replay the recorded trajectories in simulation, tracking joint angles with
a high-gain PD controller. We use hydroelastic contact with high moduli
($10^8\text{~Pa}$) to model hard objects. Thin, complex collision geometries,
varying masses, high elastic moduli, heavy robot arms, and stiff controller
gains make this scenario particularly challenging. 

\begin{figure*}
    \centering
    \begin{subfigure}{0.19\linewidth}
        \centering
        \includegraphics[width=\linewidth]{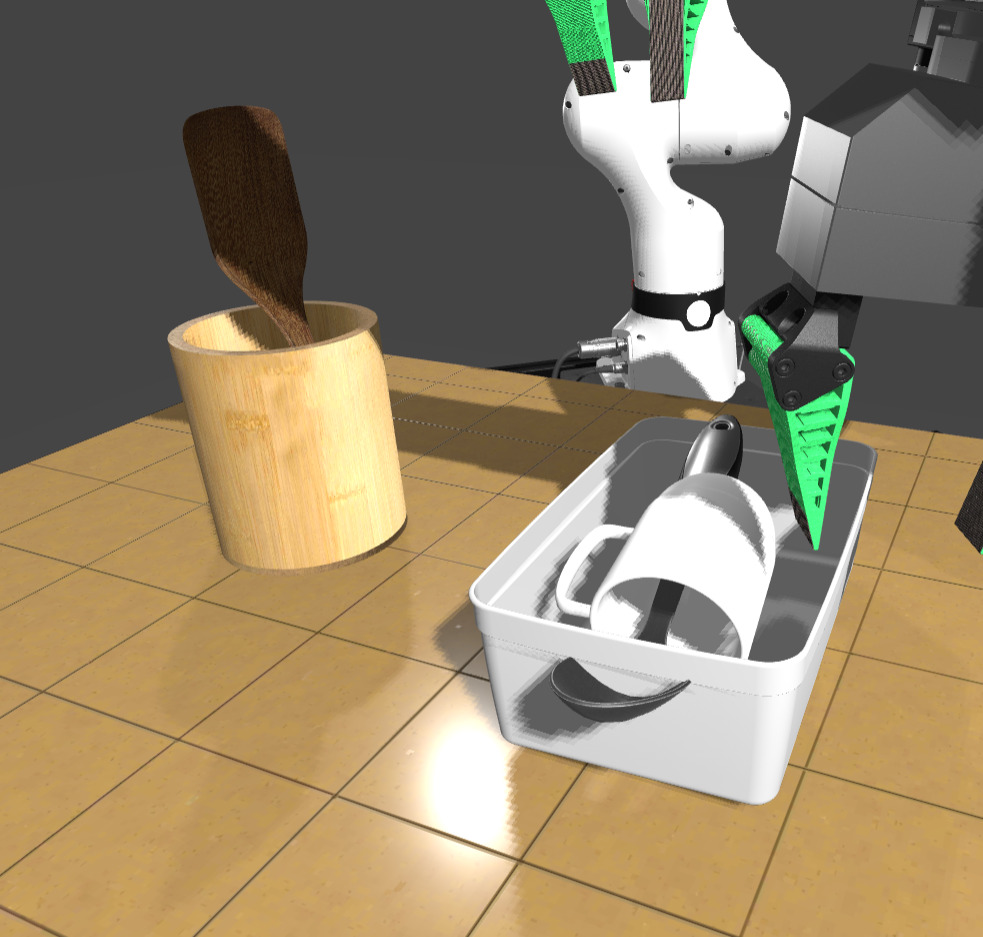}
        \caption{Fixed step, $\delta t = 0.02$ s}
        \label{fig:artifacts:fixed_20ms}
    \end{subfigure}
    \begin{subfigure}{0.19\linewidth}
        \centering
        \includegraphics[width=\linewidth]{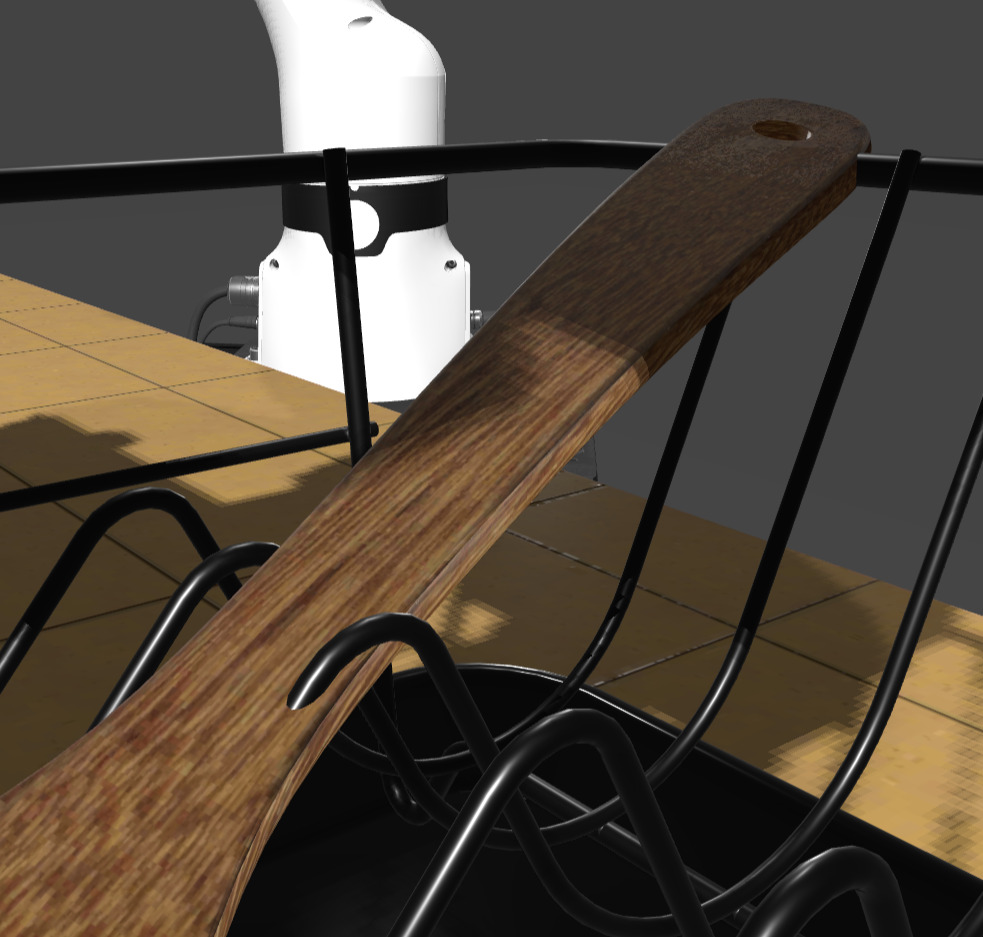}
        \caption{Fixed step, $\delta t = 0.01$ s}
        \label{fig:artifacts:fixed_10ms}
    \end{subfigure}
    \begin{subfigure}{0.19\linewidth}
        \centering
        \includegraphics[width=\linewidth]{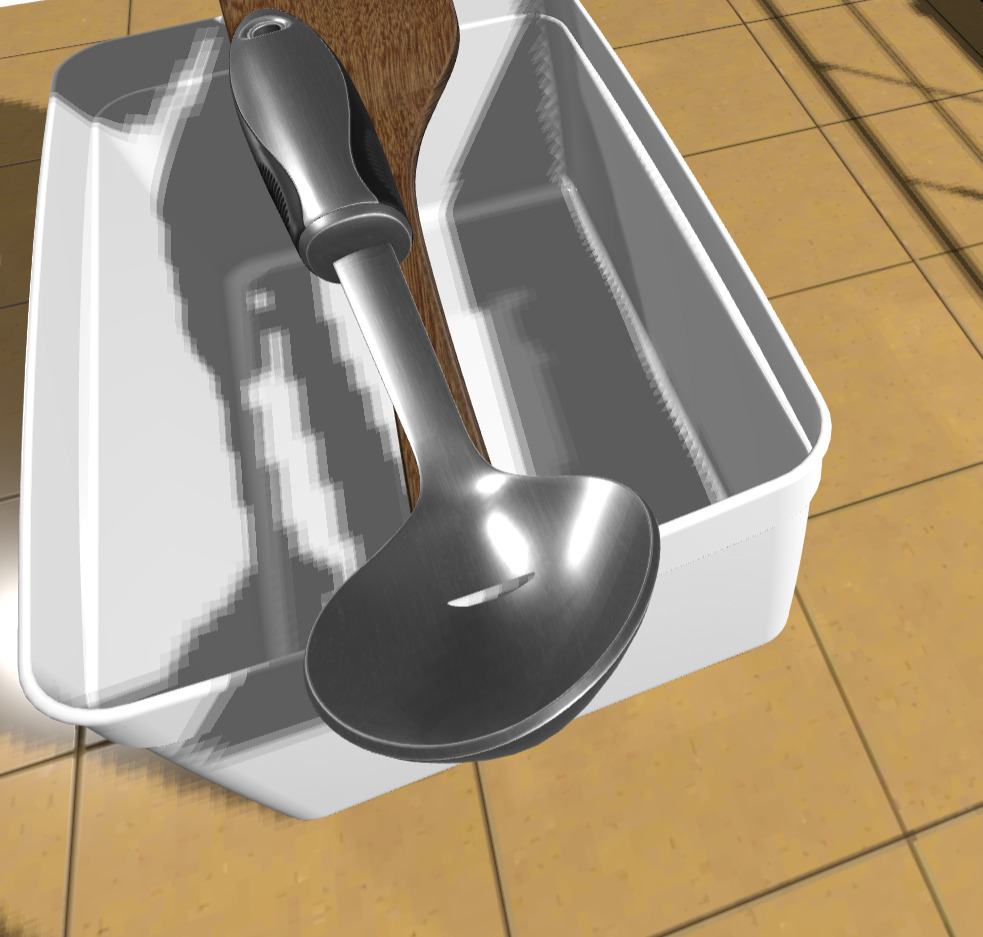}
        \caption{Fixed step, $\delta t = 0.002$ s}
        \label{fig:artifacts:fixed_2ms}
    \end{subfigure}
    \begin{subfigure}{0.19\linewidth}
        \centering
        \includegraphics[width=\linewidth]{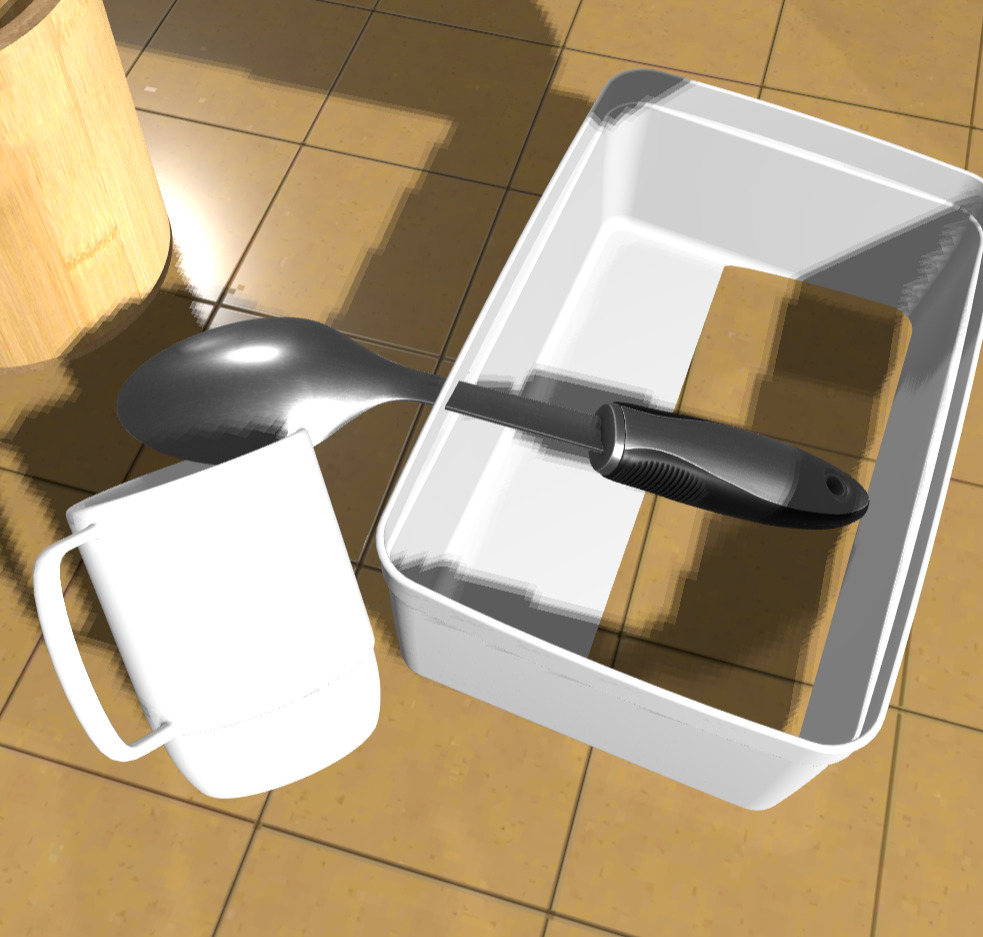}
        \caption{Error control, $\varepsilon_\text{acc} = 10^{-1}$}
        \label{fig:artifacts:error_control_01}
    \end{subfigure}
    \begin{subfigure}{0.19\linewidth}
        \centering
        \includegraphics[width=\linewidth]{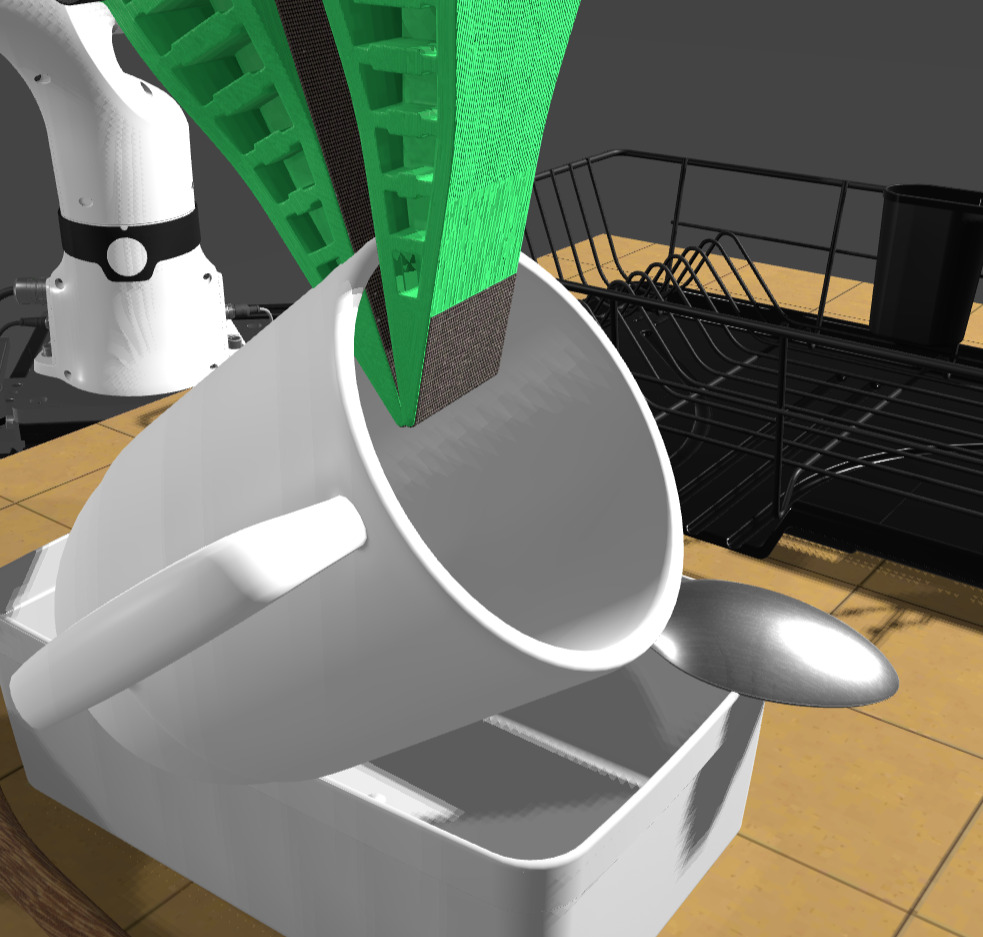}
        \caption{Error control, $\varepsilon_\text{acc} = 10^{-2}$}
        \label{fig:artifacts:error_control_001}
    \end{subfigure}
    \caption{Undesirable artifacts from simulating the dish rack example
    (Fig.~\ref{fig:hero}) with a too-large time step (fixed-step mode) or a
    too-loose accuracy (error control mode). Tightening the accuracy is a more
    efficient means of eliminating these artifacts than reducing the time step,
    as shown in Table~\ref{tab:hero_demo_artifacts}
    }
    \label{fig:artifacts}
\end{figure*}

Discrete time-stepping struggles with major artifacts on this task, as shown in
Fig.~\ref{fig:artifacts}. Thin objects interacting with the bin and the wire
dishrack make the simulation vulnerable to passthrough and rattling, despite the
use of a sophisticated hydroelastic contact model
\cite{elandt2019pressure,masterjohn2022velocity}. As shown in
Table~\ref{tab:hero_demo_artifacts}, we need to reduce the time step to 1~ms to
eliminate all visually-obvious artifacts. With this small time step, the
simulation runs much slower than real time.

\begin{table}
    \centering
    \begin{tabular}{|c|c|c|c|c|}
        \hline
        Accuracy & $10^{-1}$ & $10^{-2}$ & $\mathbf{10^{-3}}$ & $\mathbf{10^{-4}}$ \\
        \hline
        Real-time rate & 218\% & 463\% & \textbf{342\%} & \textbf{132\%}\\
        \hline
        Artifacts & Yes (\ref{fig:artifacts:error_control_01}) & Yes (\ref{fig:artifacts:error_control_001}) & \textbf{No} & \textbf{No} \\
        \hline
        \hline
        Time Step (s) & $0.02$ & $0.01$ & $0.002$ & $\mathbf{0.001}$ \\
        \hline
        Real-time rate &102\% & 180\%& 43\% & \textbf{21\%}\\
        \hline
        Artifacts & Yes (\ref{fig:artifacts:fixed_20ms}) & Yes (\ref{fig:artifacts:fixed_10ms}) & Yes (\ref{fig:artifacts:fixed_2ms}) & \textbf{No} \\
        \hline
    \end{tabular}
    \caption{Performance comparison between error-controlled integration (top)
    and fixed-step integration (bottom) on the dishrack simulation
    (Fig.~\ref{fig:hero}). Error-controlled integration eliminates obvious
    artifacts (shown in Fig.~\ref{fig:artifacts}) while achieving faster overall
    simulation speeds.}
    \label{tab:hero_demo_artifacts}
\end{table}

\begin{figure*}
    \centering
    \includegraphics[width=0.19\linewidth]{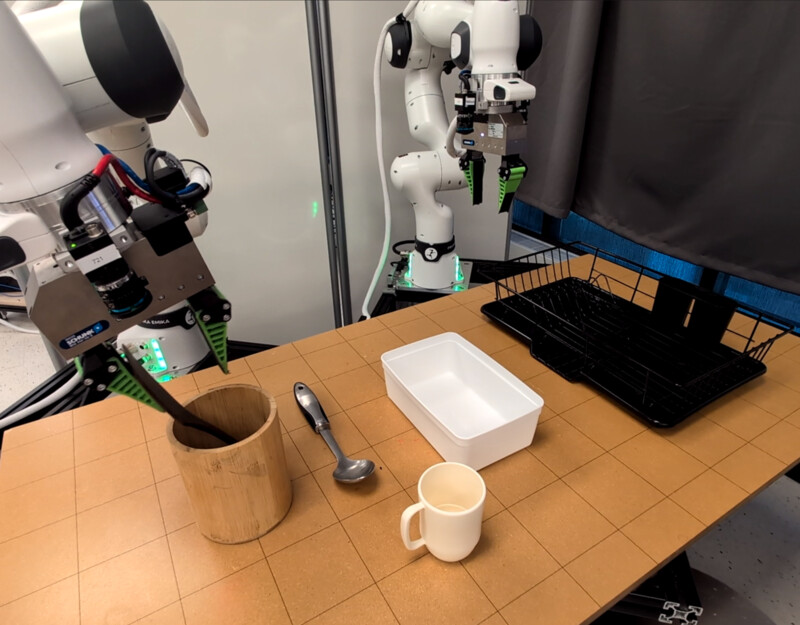}
    \includegraphics[width=0.19\linewidth]{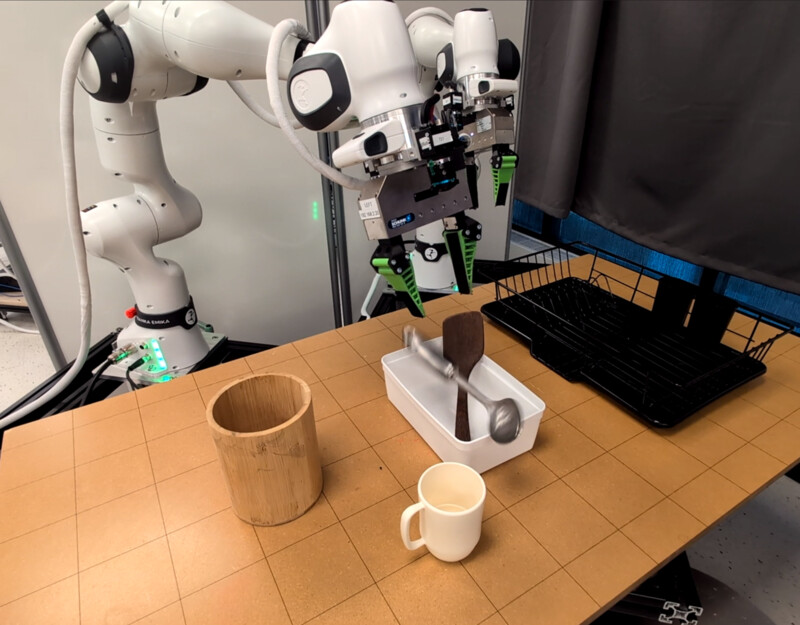}
    \includegraphics[width=0.19\linewidth]{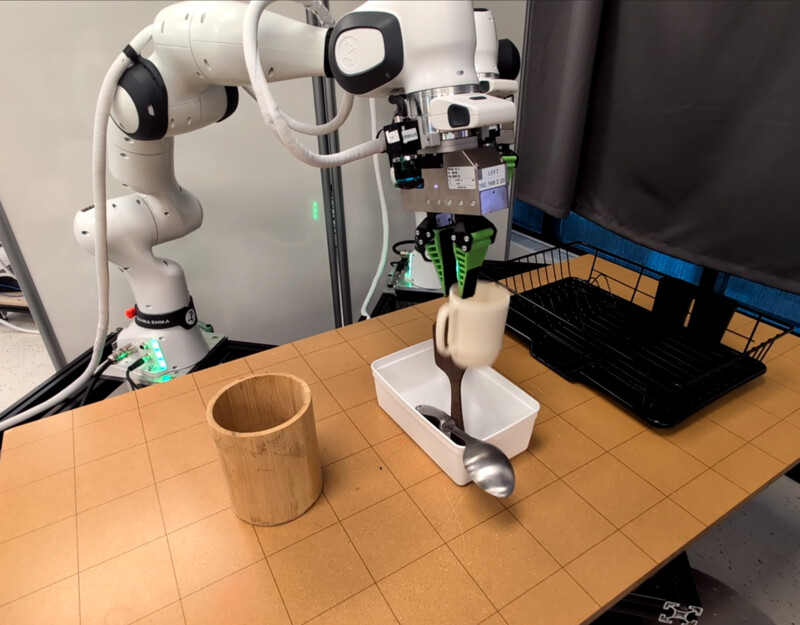}
    \includegraphics[width=0.19\linewidth]{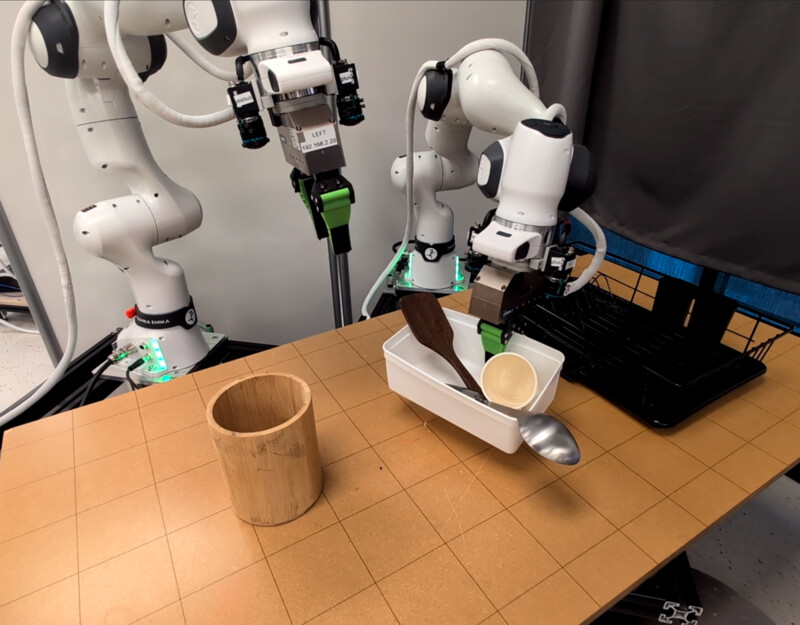}
    \includegraphics[width=0.19\linewidth]{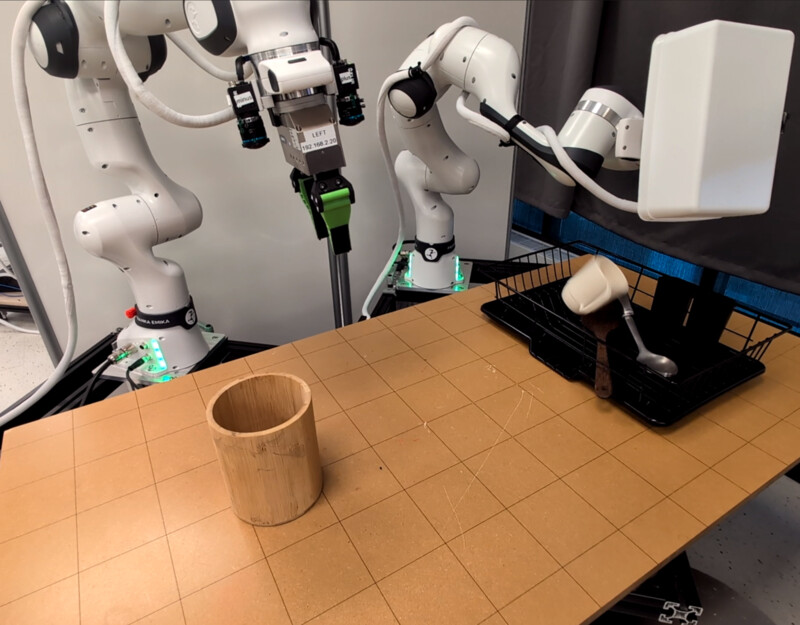}
    
    \includegraphics[width=0.19\linewidth]{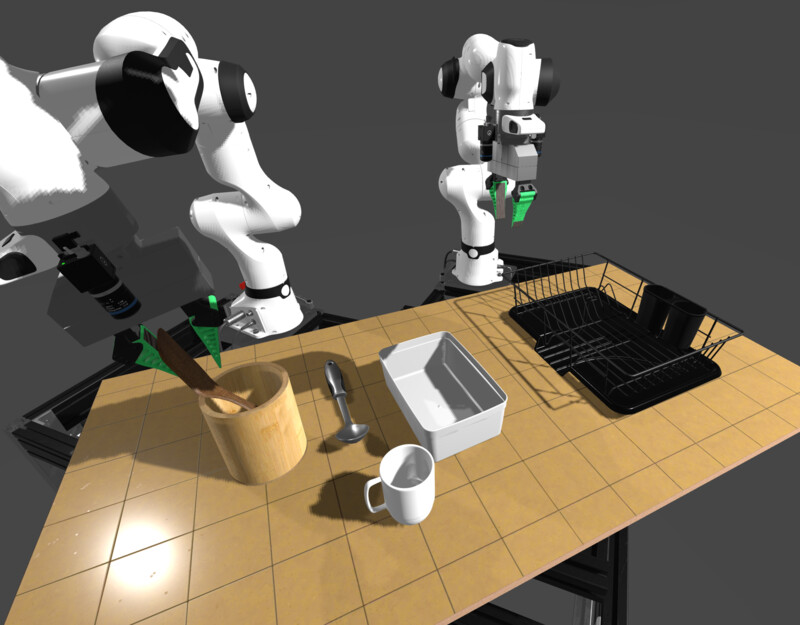}
    \includegraphics[width=0.19\linewidth]{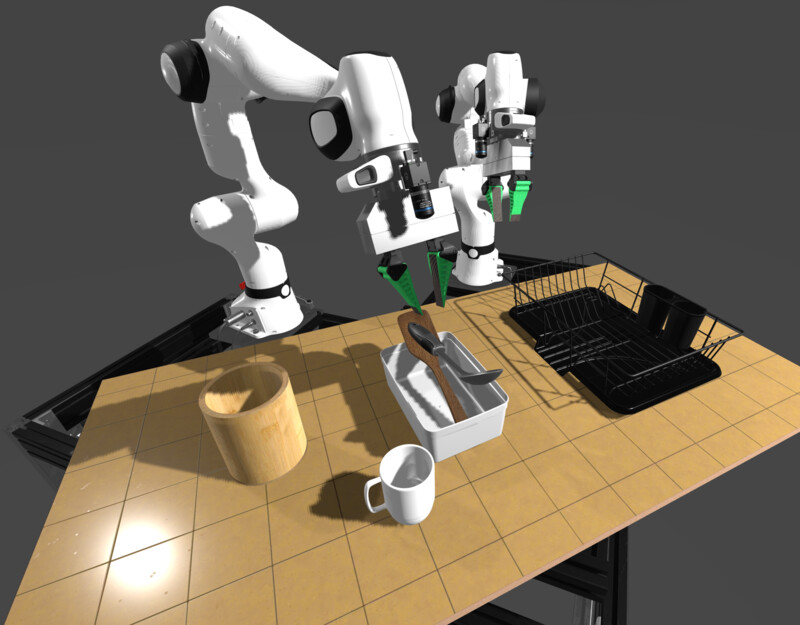}
    \includegraphics[width=0.19\linewidth]{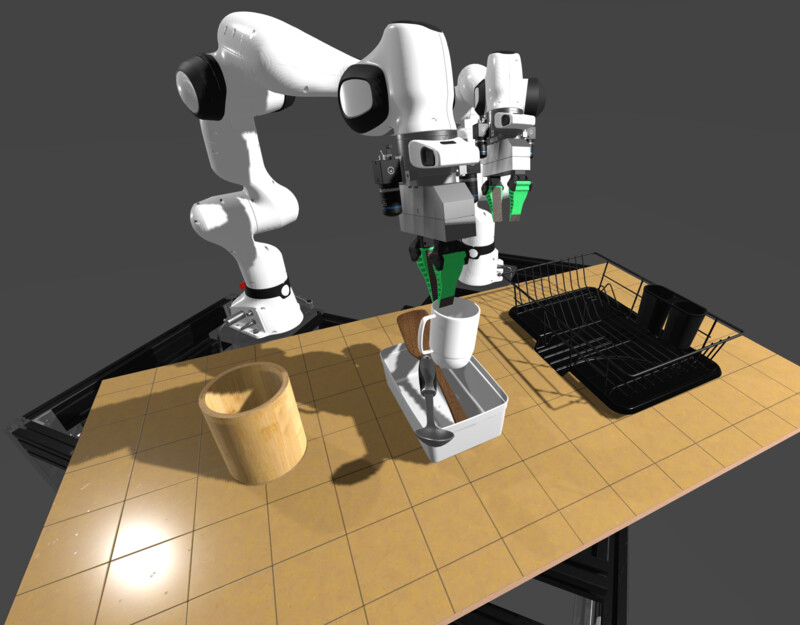}
    \includegraphics[width=0.19\linewidth]{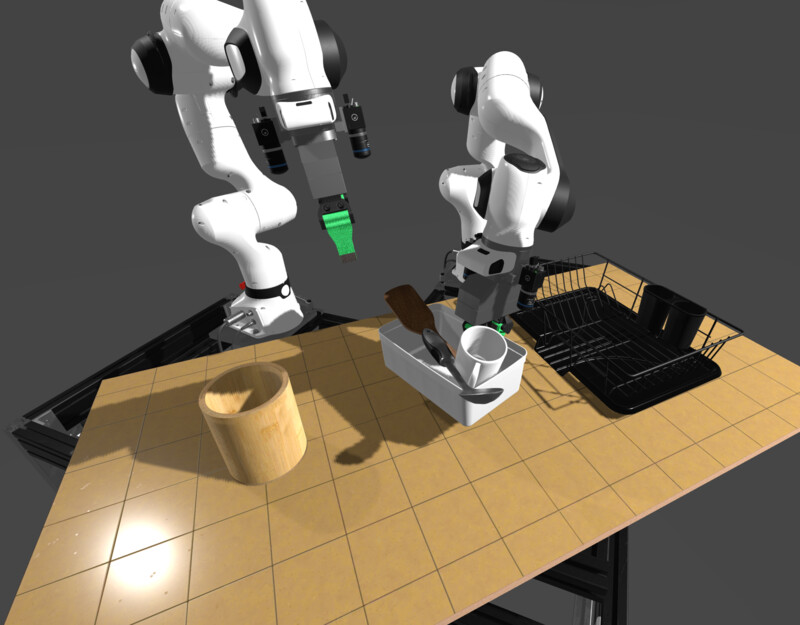}
    \includegraphics[width=0.19\linewidth]{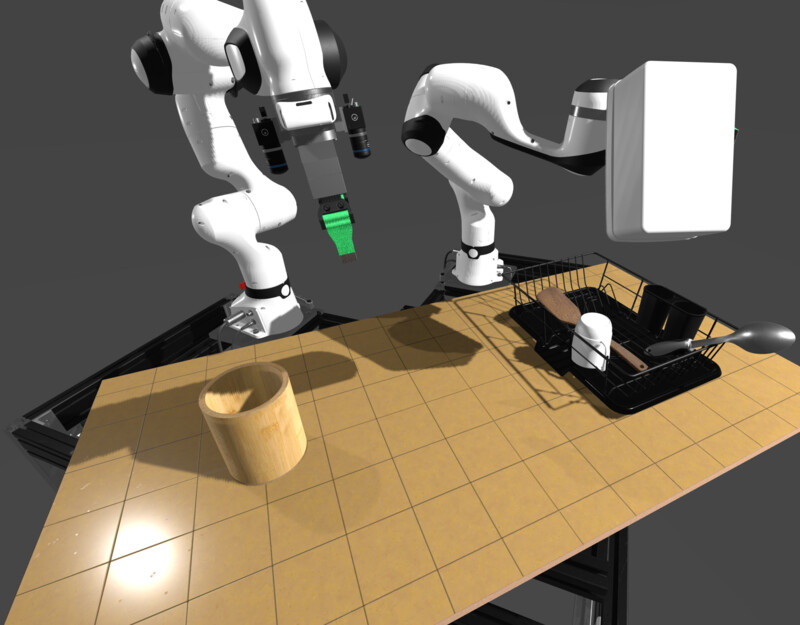}

    \caption{Snapshots of the dish rack task on hardware (top) and in simulation
    with CENIC (bottom, $\varepsilon_\text{acc} =
    10^{-3}$). The simulation and real trajectories are not exactly the same,
    due to model sensitivity and chaotic dynamics. But the overall qualitative
    behavior in simulation matches the real-world demonstration: there are no
    drops, missed grasps, passthrough incidents, rattling, or other major
    numerical artifacts. In contrast, standard discrete time-stepping methods
    introduce major artifacts, even with relatively small time steps
    (Fig.~\ref{fig:artifacts:fixed_20ms}-\subref{fig:artifacts:fixed_2ms}).}
    \label{fig:dishrack_screenshots}
\end{figure*}

In contrast, error-controlled CENIC eliminates all visually-obvious artifacts at
an accuracy of $\varepsilon_\text{acc} = 10^{-3}$, while completing the
simulation faster than real time (342\% real-time rate). Screenshots comparing
the simulation and real-world trajectories are shown in
Fig.~\ref{fig:dishrack_screenshots}. While the trajectories do not match exactly
(unsurprising given chaotic dynamics and open-loop playback), the results are
qualitatively similar and obvious discretization artifacts are absent. Moreover,
contact forces are smooth, leading to robust grasping.

\begin{figure}
    \centering
    \includegraphics[width=\linewidth]{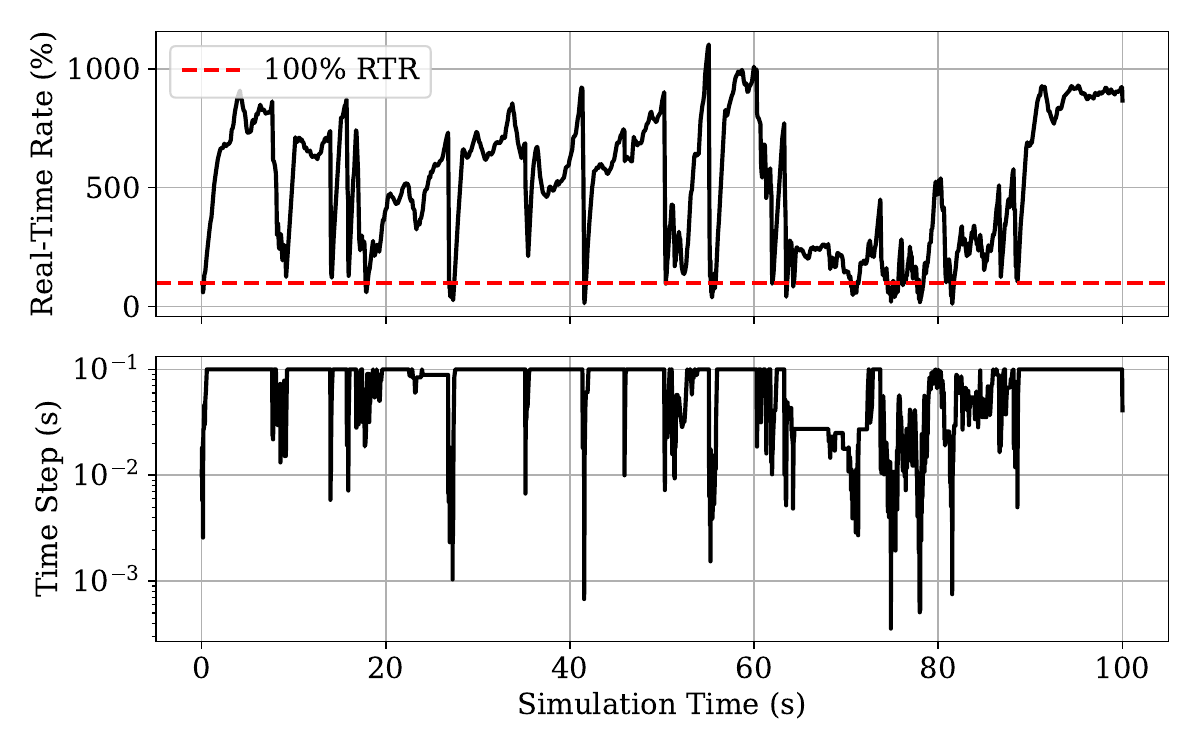}
    \caption{Real-time rate and step size $\delta t$ for the complex dishrack
    simulation (Fig.~\ref{fig:hero}) at accuracy $\varepsilon_\text{acc} =
    10^{-3}$. By only taking small time steps when needed to resolve stiff
    collision dynamics, CENIC achieves faster overall simulation times than
    discrete time-stepping (Table~\ref{tab:hero_demo_artifacts}).}
    \label{fig:hero_demo_realtime_rate}
\end{figure}

With error control, the real-time rate (RTR) varies throughout the simulation,
as shown in Fig.~\ref{fig:hero_demo_realtime_rate}. \hl{RTR is defined as the
ratio of simulated time to wall time; RTR $> 100\%$ indicates that the
simulation is running faster than real time.} During complex collisions, such as
when the robot dumps objects onto the wire rack around 70--80~s, the time step
decreases, reducing the RTR. As the objects settle, the simulator takes larger
steps. \textbf{This adaptive stepping enables faster overall simulation despite
the added cost of error estimation.} \hl{This advantage is thanks in large part
to the convexity of the underlying solve. Non-convex solvers are rarely able to
ensure convergence with time steps as large as 0.1~s, even with advanced
features like transition-aware linesearch \cite{castro2020transition}.}

\subsection{Performance Profiling}\label{sec:experiments:performance_profiling}

In this section, we characterize the computational cost breakdown of CENIC, with
particular attention to the impact of the performance optimizations described in
Section~\ref{sec:performance}.

\begin{figure}
    \centering
    \includegraphics[width=\linewidth]{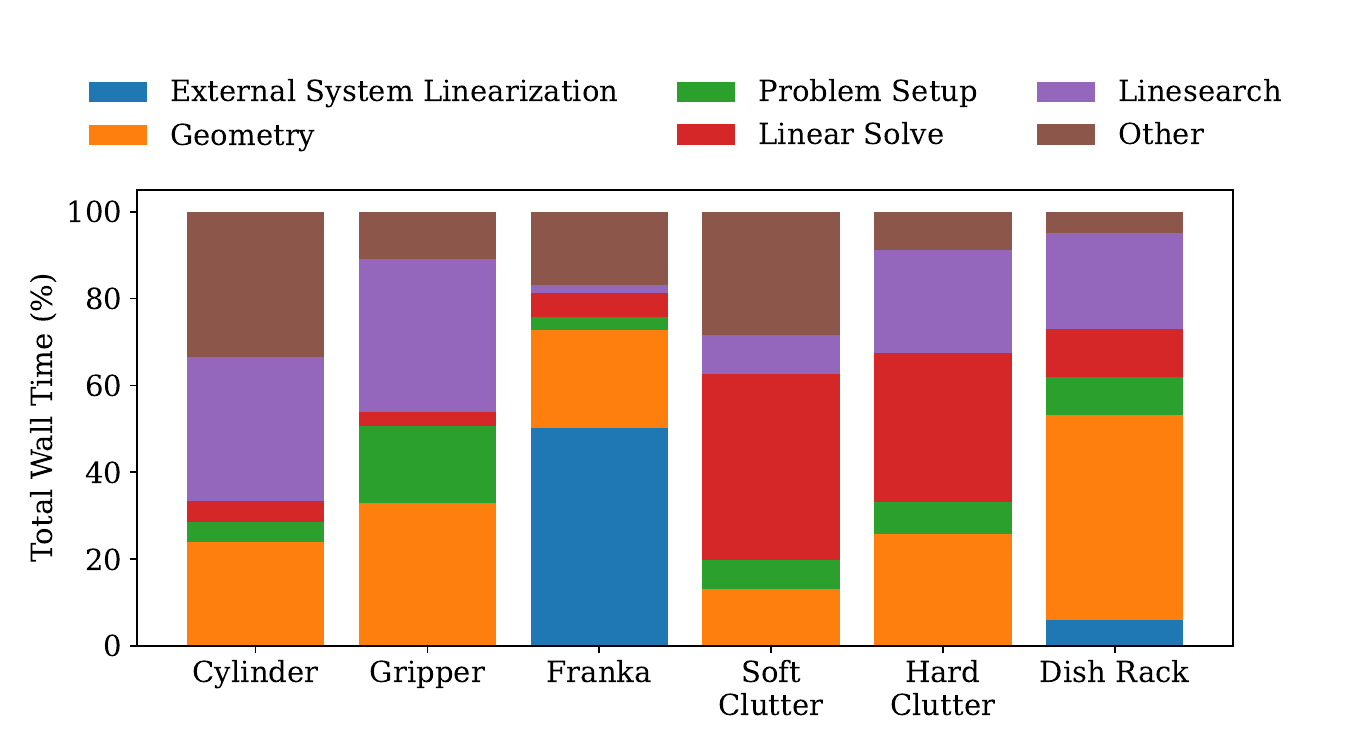}
    \caption{Computational cost breakdown for each of the examples, run with
    error control at accuracy $\varepsilon_\text{acc} = 10^{-3}$.}
    \label{fig:performance_profiling}
\end{figure}

Figure~\ref{fig:performance_profiling} shows the computational cost of major
components for our five main examples (Fig.~\ref{fig:examples}) as well as the
dish rack demonstration (Fig.~\ref{fig:dishrack_screenshots}), all at accuracy
$\varepsilon_\text{acc} = 10^{-3}$. The breakdown varies considerably between
problems, with no single component always dominating. Geometry queries are often
a major component, particularly for scenarios with hydroelastic contact
(cylinder, gripper, franka, dish rack).

Table~\ref{tab:adaptive_convergence} compares performance with and without the
adaptive convergence criterion from Sec.~\ref{sec:performance:convergence}. The
adaptive criterion, which uses looser tolerances under looser accuracies,
enables mild speedups. While a larger $\kappa$ can produce faster solve times,
we found that a more conservative $\kappa = 10^{-3}$ provides a good balance
between speed and stable resolution of contact forces.

\begin{table}
    \centering
    \begin{tabular}{|c|c|c|c|c|c|}
        \hline
        Accuracy & $10^{-1}$ & $10^{-2}$ & $10^{-3}$ & $10^{-4}$ & $10^{-5}$ \\
        \hline
        Wall time (adaptive) & 0.19 & 0.39 & 0.65 & 1.11 & 2.69 \\
        \hline
        Wall time (fixed) & 0.21 & 0.41 & 0.71 & 1.19 & 2.69 \\
        \hline
        Iterations (adaptive) & 4015 & 9973 & 19235 & 39537 & 95839 \\
        \hline
        Iterations (fixed) & 4591 & 10938 & 20865 & 40729 & 95839 \\
        \hline
    \end{tabular}
    \caption{Impact of adaptive convergence criteria (Sec~\ref{sec:performance:convergence}) on hard clutter simulation
        times (in seconds) and total solver iterations. }

    \label{tab:adaptive_convergence}
\end{table}
        
Table~\ref{tab:hessian_reuse} characterizes the impact of Hessian reuse
(Sec.~\ref{sec:performance:hessian_reuse}) on the hard clutter example. While
similar Jacobian reuse strategies are essential for obtaining good performance
in implicit Runge--Kutta schemes \cite{hairer1996solving}, we find that Hessian
reuse provides only a modest benefit with error control enabled, and no
improvement in fixed-step mode \hl{for this particular example. In fixed step
mode, we see that Hessian reuse does reduce the number of factorizations, but
this benefit is outweighed by the need for more iterations. While Hessian reuse
may provide some benefit in fixed-step mode in other scenarios, it tends to be
more \hll{beneficial} with error-control because our error-control strategy
requires several closely related convex solves at each step
(Sec.~\ref{sec:main:step_doubling}).} 

\begin{table}
    \centering
    \begin{tabular}{|c|c|c|c|c|c|c|}
        \hline
        Error control & Reuse & Wall time (s) & Steps & Facts. & Iters. \\
        \hline
        Off & Off & 0.26 & 1000 & 3980 & 3980 \\
        Off & On & 0.27 & 1000 & 2293 & 6068 \\
        \hline
        On & Off & 0.65 & 499 & 16624 & 16624 \\
        On & On & 0.61 & 510 & 10557 & 19235 \\
        \hline
    \end{tabular}
    \caption{Impact of Hessian reuse on the hard clutter example. Hessian reuse
        reduces the number of expensive factorizations (``Facts."), at the cost
        of more iterations. With error control enabled ($\varepsilon_{\text{acc}} =
        10^{-3}$, maximum timestep 0.1~s), this offers a small improvement in
        overall performance. In fixed step mode ($\delta t = 10$~ms), Hessian
        reuse does not improve performance.}
    \label{tab:hessian_reuse}
\end{table}

\begin{figure}
    \centering
    \includegraphics[width=\linewidth]{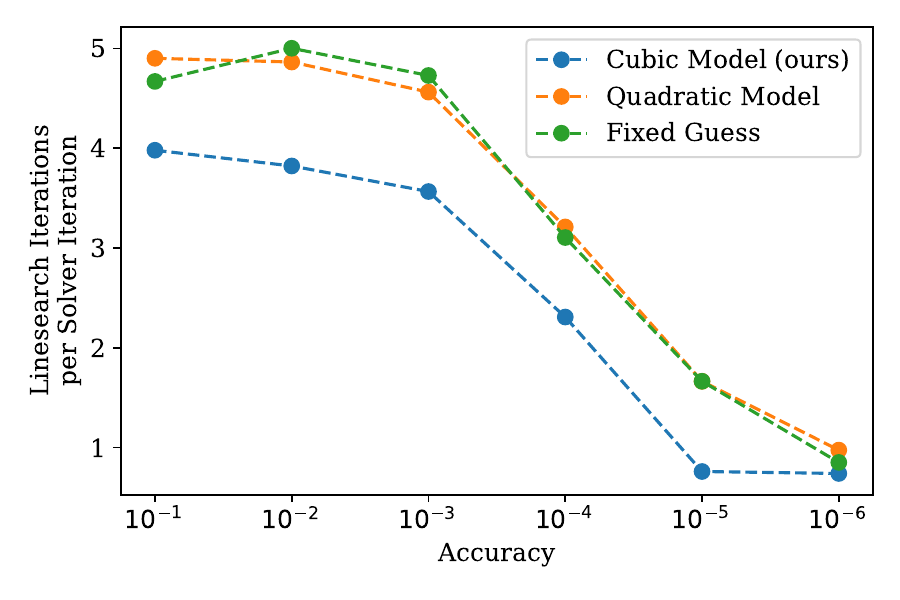}
    \caption{Comparing different strategies for choosing an initial guess for
        linesearch, for the hard clutter problem at different accuracies. Our
        cubic spline method consistently reduces the number of iterations
        required for exact linesearch, resulting in modest improvements in
        overall speed. }
    \label{fig:linesearch_iteration}
\end{figure}

Figure~\ref{fig:linesearch_iteration} quantifies the impact of cubic linesearch
initialization (Sec.~\ref{sec:performance:linesearch}). Note that as accuracy is
tightened, fewer linesearch iterations are required, since the optimization
problem is easier to solve for small time steps. Cubic linesearch initialization
offers a small but consistent reduction in the total number of linesearch
iterations relative to a fixed-guess strategy ($\alpha_0 = 1.0$) and the
quadratic initialization strategy implemented in \cite{castro2022unconstrained}.

\hl{Finally, Table~\ref{tab:stiffness_comparison} characterizes CENIC's
performance under a range of contact stiffnesses. As stiffness increases, the
problem becomes more difficult, generally requiring more time steps and solver
iterations (modulo some variation due to the chaotic nature of the system).
Nonetheless, CENIC maintains strong, faster-than-real-time performance even with
contact stiffnesses orders of magnitude greater than steel.}

\begin{table}
    \centering
    \hl{
    \begin{tabular}{|c|c|c|c|}
        \hline
        Contact Stiffness (N/m) & Steps & Iterations & Wall time (s) \\
        \hline
        $10^2$ & 585 & 2232 & 0.22 \\
        $10^4$ & 440 & 11179 & 0.34 \\
        $10^6$ & 664 & 24765 & 0.92 \\
        $10^8$ & 1378 & 37183 & 1.742 \\
        $10^{10}$ & 1621 & 34647 & 1.53 \\
        $10^{12}$ & 1641 & 37495 & 1.63 \\
        $10^{14}$ & 1865 & 39009 & 1.98 \\
        $10^{16}$ & 2140 & 60761 & 2.83 \\
        \hline
    \end{tabular}
    }
    \caption{\hl{Statistics for hard clutter simulation (10 seconds of
    simulation time) at accuracy $\varepsilon_\text{acc} = 10^{-3}$, and varying
    contact stiffness $k_i$. Stiffer contact parameters lead to more time steps,
    more iterations, and longer wall times. Nonetheless, CENIC is able to
    simulate this contact-rich scene significantly faster than real-time even
    with extremely stiff contact parameters (note that steel has stiffness
    around $k_i = 10^7$ N/m \cite{castro2024irrotational}).}}
    \label{tab:stiffness_comparison}
\end{table}
\section{Discussion and Limitations}\label{sec:discussion}

CENIC is the first error-controlled integrator specialized for robotics,
outperforms existing error-controlled integrators by orders of magnitude, and
runs at speeds comparable to state-of-the-art discrete-time simulators. 

Error control yields inherent robustness: users can focus on modeling their
controllers and external systems, while CENIC automatically adapts the time step
to ensure accuracy and stabilize stiff dynamics that would otherwise become
unstable under standard discrete schemes. With CENIC, users specify a desired
accuracy rather than a time step, eliminating a common pain point in authoring
multibody simulations. Moreover, by controlling accuracy, CENIC separates the
numerical truncation error introduced by the time-stepping scheme from modeling
error arising from simplifying mathematical assumptions, inaccurate parameters,
or user mistakes.

CENIC incorporates engineering-grade, experimentally validated contact models
and, for the first time, provides a formal framework with mathematical
guarantees for modeling dry friction using distinct static and dynamic
coefficients while meeting the high speed requirements of robotics simulation.
This stands in contrast to common game-engine approaches, where PGS-style
solvers handle friction iteratively through ad-hoc clipping heuristics, which
lack convergence and accuracy guarantees and can produce artifacts or
difficult-to-diagnose instabilities.

Thanks to its basis in convex ICF, CENIC guarantees convergence at each time
step. At the same time, CENIC maintains consistency: numerical artifacts vanish
as accuracy is tightened, recovering the true trajectory of a given model.

\hl{Like many convex simulators
\cite{castro2022unconstrained,castro2024irrotational,todorov2012mujoco}, CENIC
uses Newton's method for convex optimization. Exploring alternative optimization
algorithms, such as alternating direction method of multipliers (ADMM) and
conjugate gradient (CG) methods remains an interesting avenue for future
research, particularly as error control relaxes the need for extremely tight
convergence (Sec.~\ref{sec:performance:convergence}). }

For interactive simulation, \hl{our experience is} that CENIC is most effective
at loose accuracy settings around $\varepsilon_\text{acc} = 10^{-3}$.
\hl{Empirically, this was sufficient} to eliminate major discretization
artifacts in dexterous manipulation scenarios. 

\hl{CENIC does not guarantee non-interpenetration for thin objects. However,
tightening the requested accuracy tends to eliminate such artifacts. Since
$\varepsilon_\text{acc}$ can be roughly interpreted as a length scale, we
recommend simulating with an accuracy slightly smaller than the smallest feature
of interest in the scene, and adjusting as needed to obtain good performance.}

Since step-doubling is first order, it is less efficient at tight tolerances,
where higher-order schemes may offer better performance, an interesting avenue
for future research. Notably, we observed that instability in a second-order
trapezoidal method severely degraded performance compared to first-order
step-doubling.

\hl{CENIC brings the well-established advantages of variable-step
error-controlled integration to robotics simulation, overcoming the typical
limitations imposed by stiff contact dynamics. } CENIC is most effective in
offline simulation, where over long horizons it surpasses fixed-step integration
by taking larger steps during slow dynamics while maintaining accuracy and
robustness. 

\hl{Real-time rate} fluctuations currently limit \hl{CENIC's use in real-time
applications}, though we expect this to improve with CPU and GPU parallelization
(all results so far used a single CPU thread). \hl{While CENIC cannot guarantee
any particular real-time rate, we find that in practice it achieves simulation
speeds on par with state-of-the-art robotics simulators, and orders of magnitude
faster than other \hll{error-controlled} integration schemes. This indicates
strong potential for real-time applications like teleoperation and
model-predictive control. Recent work combining off-the-shelf error-controlled
integrators with MuJoCo MJX \cite{paulusdifferentiable} further supports the
promise of hardware-accelerated error-controlled simulation for learning,
optimization, and control applications.}

We believe that fast error-controlled robotics simulation will open the door to
better sim-to-real transfer, model identification, policy learning, and more.
Future work will focus on verifying the usefulness of CENIC in these
applications, with a particular focus on systematic hardware validation of
sim-to-real transfer.

\section{Conclusion}\label{sec:conclusion}

We presented CENIC, a new method for simulating multibody dynamics with contact
that combines modern convex time-stepping with traditional error-controlled
integration. CENIC eliminates discretization artifacts, outperforms existing
error-controlled integrators by orders of magnitude, and achieves simulation
speeds on par with state-of-the-art discrete-time robotics simulators.
Additionally, CENIC can integrate user-defined controllers implicitly and
rigorously simulate dry friction with different static and dynamic coefficients,
all while maintaining guaranteed accuracy and convergence.

Most importantly, CENIC disentangles numerical discretization error from
modeling error while maintaining the fast simulation speeds required by modern
robotics workflows. We hope that this will enable accelerated progress in
sim-to-real transfer, model identification, policy learning and evaluation, and
model-based control for robotics.

\section*{Acknowledgements}

Thanks to Mariah Smith-Jones, Allison Henry, and Phoebe Horgan for helping with
the real-to-sim experiments, the TRI Dynamics \& Simulation team for many
helpful discussions, and the TRI Robotics and Large Behavior Model divisions for
supporting this research.

\balance
\bibliographystyle{IEEEtran}
\bibliography{references}

\end{document}